\RequirePackage{fix-cm}
\documentclass[twocolumn]{svjour3}
\smartqed
\usepackage[detect-all,per-mode=reciprocal,load-configurations="binary",inter-unit-separator={},binary-units=true]{siunitx}
\usepackage{tabularx}
\usepackage{subfig}
\usepackage{graphicx}
\graphicspath{
	{../}
}

\journalname{Neural Computing and Applications}

\usepackage{amsmath}
\usepackage[misc]{ifsym}
\usepackage{array}
\usepackage[numbers]{natbib}
\usepackage{xcolor}
\usepackage{xspace}
\usepackage{amssymb}
\usepackage{dsfont}
\usepackage{enumitem}
\usepackage{booktabs}
\usepackage{hyphenat}
\usepackage{acro}
\DeclareAcronym{ABFT}{
	short = ABFT ,
	long = algorithm-based fault tolerance ,
	class = abbrev
}
\DeclareAcronym{ADCR}{
	short = ADCR ,
	long = accumulated data-computation ratio ,
	class = abbrev
}
\DeclareAcronym{ASI}{
	short = ASI , 
	long = architecture sensitivity index ,
	class = abbrev
}
\DeclareAcronym{AV}{
	short = AV , 
	long = autonomous vehicle ,
	class = abbrev
}
\DeclareAcronym{BER}{
	short = BER , 
	long = bit error rate ,
	class = abbrev
}
\DeclareAcronym{CCR}{
	short = CCR ,
	long = classification change rate ,
	class = abbrev
}
\DeclareAcronym{ConvNet}{
	short = CNN ,
	long = convolutional neural network ,
	class = abbrev
}
\DeclareAcronym{DNN}{
	short = DNN ,
	long  = deep neural network ,
	class = abbrev
}
\DeclareAcronym{DRAM}{
	short = DRAM ,
	long = dynamic random-access memory ,
	class = abbrev
}
\DeclareAcronym{GPU}{
	short = GPU ,
	long  = graphics processing unit ,
	class = abbrev
}
\DeclareAcronym{GTSRB}{
	short = GTSRB ,
	long = German Traffic Sign Recognition Benchmark ,
	class = abbrev
}
\DeclareAcronym{LRP}{
	short = LRP ,
	long  = layer-wise relevance propagation ,
	class = abbrev
}
\DeclareAcronym{LSB}{
	short = LSB,
	long = least significant bit,
	class = abbrev
}
\DeclareAcronym{MAC}{
	short = MAC ,
	long  = multiply-accumulate ,
	class = abbrev
}
\DeclareAcronym{MSB}{
	short = MSB ,
	long  = most significant bit ,
	class = abbrev
}
\DeclareAcronym{NAS}{
	short = NAS ,
	long  = neural architecture search ,
	class = abbrev
}
\DeclareAcronym{RAM}{
	short = RAM ,
	long  = random-access memory ,
	class = abbrev
}
\DeclareAcronym{ReLU}{
	short = ReLU ,
	long  = rectified linear unit ,
	class = abbrev
}
\DeclareAcronym{SDC}{
	short = SDC ,
	long = silent data corruption ,
	class = abbrev
}
\DeclareAcronym{SGD}{
	short = SGD ,
	long = stochastic gradient descent ,
	class = abbrev
}
\DeclareAcronym{SRAM}{
	short = SRAM ,
	long = static random-access memory ,
	class = abbrev
}
\DeclareAcronym{TMR}{
	short = TMR ,
	long = triple modular redundancy ,
	class = abbrev
}

\newcommand{\pnash}{\texttt{LEMONADE}\xspace}
\newcommand{\pf}{\mathcal{P}  }
\newcommand{\pkde}{p_\mathrm{KDE}}
\newcommand{\NN}{\mathcal{N}}
\newcommand{\moaf}{\mathfrak{f}}
\newcommand{\fcheap}{f_\mathrm{cheap}}
\newcommand{\fexp}{f_\mathrm{exp}}
\newcommand{\eg}{e.g.\ }
\newcommand{\ie}{i.e.\ }
\newcommand{\etal}{et al.\ }

\begin{document}

\title{Automated design of error-resilient and hardware-efficient deep neural networks
}


\author{Christoph~Schorn\textsuperscript{1,2}	\and
        Thomas~Elsken\textsuperscript{3,4}		\and
        Sebastian~Vogel\textsuperscript{1,2}	\and
        Armin~Runge\textsuperscript{1}			\and
        Andre~Guntoro\textsuperscript{1}		\and
        Gerd~Ascheid\textsuperscript{2}
}

\authorrunning{Schorn et al.} 

\institute{
	\begin{tabular}{@{}l@{\hskip7pt}>{\raggedright\arraybackslash}p{.44\textwidth}}
		\Letter 			& Christoph Schorn\\
							& Christoph.Schorn@de.bosch.com\\
							& \\
		\textsuperscript{1} & Department of Dependable Connected Systems, Bosch Corporate Research, Renningen, Germany\\
		\textsuperscript{2} & Institute for Communication Technologies and Embedded Systems, RWTH Aachen University, Aachen, Germany\\
		\textsuperscript{3} & Bosch Center for Artificial Intelligence, Renningen, Germany\\
		\textsuperscript{4} & Department of Computer Science, University of Freiburg, Freiburg im Breisgau, Germany\\
	\end{tabular}
}


\maketitle

\begin{abstract}
Applying deep neural networks (DNNs) in mobile and safety-critical systems, such as autonomous vehicles, demands a reliable and efficient execution on hardware. Optimized dedicated hardware accelerators are being developed to achieve this. However, the design of efficient and reliable hardware has become increasingly difficult, due to the increased complexity of modern integrated circuit technology and its sensitivity against hardware faults, such as random bit-flips. It is thus desirable to exploit optimization potential for error resilience and efficiency also at the algorithmic side, \eg by optimizing the architecture of the DNN. Since there are numerous design choices for the architecture of DNNs, with partially opposing effects on the preferred characteristics (such as small error rates at low latency), multi-objective optimization strategies are necessary. In this paper, we develop an evolutionary optimization technique for the automated design of hardware-optimized DNN architectures. For this purpose, we derive a set of easily computable objective functions, which enable the fast evaluation of DNN architectures with respect to their hardware efficiency and error resilience solely based on the network topology. We observe a strong correlation between predicted error resilience and actual measurements obtained from fault injection simulations. Furthermore, we analyze two different quantization schemes for efficient DNN computation and find significant differences regarding their effect on error resilience.
\end{abstract}

\section{Introduction}
\label{sec:introduction}

The application of \acp{DNN} in safety-critical perception systems, for example \acp{AV}, poses some challenges on the design of the underlying hardware platforms. On the one hand, efficient and fast accelerators are needed, since \acp{DNN} for computer vision exhibit massive computational requirements \cite{Lin18}. On the other hand, resilience against random hardware faults has to be ensured. In many driving scenarios, entering a fail-safe state is not sufficient, but fail-operational behavior and fault tolerance are required \cite{Koop16}. However, fault tolerance techniques at the hardware level often entail large redundancy overheads in silicon area, latency, and power consumption. These overheads stand in contrast to the low-power and low-latency requirements of embedded real-time \ac{DNN} accelerators. Reliability concerns in nanoscale integrated circuits, for instance soft errors in memory and logic, represent an additional challenge for the realization of fault tolerance mechanisms at the hardware level \cite{Aitk15,Gome14,Henk13,Mutl17,Srid15}. Moreover, techniques such as near-threshold computing \cite{Dres10} and approximate computing \cite{Mitt16} are desirable to meet power constraints, but can further increase error rates.

To overcome these challenges, one option is to exploit error resilience at the algorithm level and allow for a certain degree of inaccuracy at the hardware level. This is referred to as cross-layer resilience \cite{Cart10}. Due to the implicit information redundancy of neural networks, they offer some robustness against random internal perturbations, which can be exploited in a cross-layer resilience approach. Nevertheless, error resilience is strongly influenced by the architectural design of the \ac{DNN} \cite{Scho19} as well as its internal data representations \cite{Li17}. These design choices, in turn, also influence hardware efficiency and classification performance of the network. Taking these multiple, partially opposing objectives into account in a manual \ac{DNN} design procedure is non-trivial and cumbersome.

As a solution, we develop and evaluate an efficient, automated, multi-objective \ac{NAS} technique in this paper, which holistically takes classification performance as well as hardware-specific objective functions into account. In detail, our contributions are the following:
\begin{enumerate}
	\item We derive a set of objective functions for the prediction of error resilience, energy consumption, latency and required bandwidth of \acp{DNN} on hardware, solely based on the topology of their neural architecture, allowing a fast evaluation of these objectives by avoiding the need for expensive simulations or training of the neural network.
	\item We integrate these functions in an efficient, evolutionary, multi-objective \ac{NAS} algorithm, that uses (approximate) network morphisms for a fast Pareto optimization of \acp{DNN}. 
	\item We evaluate our methods and obtained Pareto trade-offs on two popular image classification benchmarks, namely CIFAR-10 and \ac{GTSRB}. In particular, we test the predictive performance of our fast error resilience prediction metric by taking \ac{SDC} measurements, employing a memory bit-flip fault injection framework.
	\item We compare two recently introduced quantization techniques for hardware-efficient \ac{DNN} inference with respect to resulting classification performance and error resilience characteristics of the neural networks. 
\end{enumerate} 
To the best of our knowledge, this is the first paper combining error resilience and hardware efficiency optimization in the context of neural architecture search.

The remainder of this paper is structured as follows. In Section~\ref{sec:related}, we give an overview of related work. In Section~\ref{sec:hw_focused_nas}, we introduce our methodology. This includes the derivation of hardware-specific objective functions, neural network quantization techniques and the multi-objective optimization algorithm used in this paper. In Section~\ref{sec:experiments}, we evaluate the outcome of our methods on two image classification benchmarks. We analyze the trade-offs between Pareto-optimal solutions, perform fault injections to compare predicted and measured resilience, and evaluate the characteristics of two different \ac{DNN} quantization methods. We close our paper with a summary and conclusions in Section~\ref{sec:conclusions}.

\section{Background and related work}
\label{sec:related}

We now give an overview of related error resilience analysis (Section~\ref{subsec:rel:resilience}), resilience optimization techniques for neural networks (Section~\ref{sub:resilience_opt}) as well as preliminaries on multi-objective optimization (Section~\ref{sub:moo}) and \acl{NAS} (\acs{NAS}) (Section~\ref{sub:nas}).

\subsection{Neural network resilience analysis}\label{subsec:rel:resilience}
Understanding a neural network's resilience against erroneous perturbations in its internal computations has been a topic of interest for decades already. Here, we give an overview of the most recent studies that target error resilience analysis of modern \acp{DNN}. An in-depth review of previous literature has been recently given by Torres-Huitzil and Girau \cite{Torr17}. 

\subsubsection{Experimental analysis}
The majority of studies on error resilience in neural networks has been experimental. They range from physical fault induction experiments in real hardware devices \cite{Sant19,What18}, over fault injections in (virtual) hardware models \cite{Aziz18,Li17,Sala18,Sant19}, to error simulations at the algorithmic behavior level \cite{Marq17,Reag18,Scho18}. Behavioral analysis can be connected to realistic hardware faults in a second step, by mapping the effect of these faults to error models in the algorithm domain \cite{Piur01}. For the model-based analysis, stuck-at-zero, stuck-at-one and random bit-flips of memory cells are commonly used. Stuck-at types are used to model permanent faults (\eg resulting from manufacturing defects) and bit-flips are typically used to model radiation-induced transient faults that lead to soft errors \cite{Torr17}. 

In summary, experimental studies found different determinants of neural network resilience, the most important being the number and type of errors, the data representation of the neural network, the \ac{DNN} type and the location where the error occurs. However, while experimental evaluation is useful for an accurate a posteriori resilience determination of a given \ac{DNN} on hardware, it is cumbersome and provides only limited insight into a priori design choices for \ac{DNN} developers to improve resilience at the algorithm level.

\subsubsection{Theoretical analysis}
A theory-guided resilience analysis offers the advantage of being more directly interpretable and avoids lengthy fault injection experiments. El Mhamdi and Guerraoui \cite{ElM17} analytically derived easily computable bounds for the forward error propagation of neurons that are stuck-at-zero (crashed neurons) and for neurons that transmit arbitrary values (Byzantine neurons). They found that the choice of activation function and number of neurons per layer are design choices that affect the forward error propagation. More precisely, an activation function with a low Lipschitz constant as well as a high number of neurons per layer can reduce forward error propagation.

A different analytical technique to derive neuron resilience prediction has been used in the context of approximate neural network computing. Backpropagation of error gradients, comparable to the technique used to determine weight updates during neural network training, has been used to estimate the average output sensitivity to perturbations in individual neurons \cite{Venk14,Zhan15}. 

Recently, Schorn \etal \cite{Scho18} showed that a technique based on \ac{LRP} \cite{Bach15} outperforms gradient-based resilience prediction. Contrarily to gradient methods, which determine the sensitivity to small perturbations in neurons, \ac{LRP} attributes to each neuron its absolute contribution to the \ac{DNN} output \cite{Mont18}, which can be interpreted as layerwise Taylor decomposition \cite{Mont17}. A high neuron relevance, averaged over a training set of input samples, corresponds to a high sensitivity against errors \cite{Scho18}. 

\subsection{Neural network resilience optimization}\label{sub:resilience_opt}
The optimization of neural network error resilience at the algorithm level is an active field of research. A number of publications simulate the effects of hardware faults during neural network training to improve resilience \cite{Deng15,Yang17,Kim18,Xia17,Liu17}. Reference \cite{Deng15} considers timing variations, \cite{Yang17,Kim18} static \ac{RAM} supply voltage scaling and \cite{Xia17,Liu17} hard defects in memristors and resistive \ac{RAM} respectively. The drawback of these approaches is that they complicate the training process, since fault injections have to be performed by placing hardware in the training loop or through realistic fault simulations. Common regularizing techniques, such as dropout \cite{Kerl93,Sriv14} and weight decay \cite{Krog91}, also improve the general error resilience of neurons \cite{ElM17}.

A second approach is to adjust the mapping of the algorithm to hardware for an optimized resilience. A significance-driven mapping of network weight bits to memory cells with different resilience has been suggested in \cite{Srin16}. However, the authors did not follow an analytical approach to determine weight resiliencies, but relied on their experience. In contrast, the \ac{LRP}-based method in \cite{Scho18} gives a theoretically founded resilience mapping of neurons.  

A third approach is to use modifications in hardware that are tailored to exploit the algorithmic resilience properties of neural networks. This can be zero-biased \cite{Aziz18} or selectively hardened \cite{Li17} memory cells, optimized data representations \cite{What18}, masking techniques \cite{Reag16,Sala18}, anomaly detectors \cite{Li17,Scho18b} and relaxed versions of classical fault tolerance mechanisms, such as \ac{TMR} \cite{Mahd12} and \ac{ABFT} checksums \cite{Sant19}.

Modifications of the neural architecture to increase resilience have been proposed as well. Dias \etal \cite{Dias10} suggest a resilience optimization procedure by replication of critical neurons and weights. However, they use exhaustive simulation to determine criticality values, which is infeasible for large-scale \acp{DNN}. Schorn et al. \cite{Scho19} showed that critical layers can be identified using \ac{LRP}. Nevertheless, no automated neural architecture design technique that jointly optimizes error resilience as well as other desirable performance and efficiency objectives of \acp{DNN} has been introduced so far. 

\subsection{Multi-objective optimization}\label{sub:moo}
In multi-objective optimization (see, \eg \citep{miettinen_evolutionary_1999}), one tries to optimize multiple, complementary objective functions $f_1,\dots, f_k$ over a space $\mathcal{N}$ of feasible solutions (in our case: a space of neural network architectures). Usually, there will be no $N^* \in \mathcal{N}$ that minimizes all objectives $f_1, \dots, f_k$ at the same time (as the objectives are complementary). Instead, there are multiple \emph{Pareto-optimal} solutions meaning that one cannot reduce any $f_i$ without increasing at least one $f_j$. Formally, we say that $N_{1}$ dominates $N_{2}$ iff $f_i(N_{1}) \leq f_i(N_{2})$ for every $i \in \{1, \dots, n\}$ and $f_j(N_{1}) < f_j(N_{2})$ for at least one $j$. $N^*$ is called Pareto-optimal iff $N^*$ is not dominated by any other $N \in \mathcal{N}$. The set of Pareto-optimal solutions is the so-called \emph{Pareto front}. Typically, multi-objective optimization can only determine a subset $\pf$ that approximates this Pareto front.

In order to rate the overall performance of a given neural network $N \in \pf$ across all objectives, the distance to the \emph{ideal point} can be used as a metric \cite{Blas08}. The (approximate) coordinates $y_i$ of the ideal point in each objective dimension $i \in \{1, \ldots, k\}$ can be determined by taking the componentwise minima of the objective functions $f_i(N)$ over the (approximated) Pareto front $\pf$ \cite{Ehrg03}:
\begin{equation}
y_i = \min_{N \in \pf} f_i(N), \quad i \in \{1, \ldots, k\}.
\end{equation}

To enhance comparability, a normalized version of the distance to the ideal point can be computed \cite{Blas08}. Therefore, the individual objective functions are first normalized
\begin{equation}
\bar{f}_i(N) = \frac{f_i(N) - \min_{N \in \pf} f_i(N)}{\max_{N \in \pf} f_i(N) - \min_{N \in \pf} f_i(N)}
\end{equation}
so that $0 \leq \bar{f}_i(N) \leq 1$. Then, a norm on the vector $\bar\moaf(N) = (\bar{f}_1(N), \ldots, \bar{f}_k(N))^\top \in \mathds{R}^k$ is computed to measure the distance of $N$ from the ideal point. 

Blasco \etal \cite{Blas08} suggest to take the infinity norm for the purpose of trade-off analysis:
\begin{equation}
\left\lVert\bar{\moaf}(N)\right\rVert_\infty = \max\left\{\bar{f}_i(N)\right\}, \quad i \in \{1, \ldots, k\}.
\end{equation}
That way, a score between $0$ and $1$ is obtained, which supplies information about the worst objective value. For example, a value of $1$ means that $N$ has the worst observed performance in at least one of the objectives. We refer to $\left\lVert\bar{\moaf}(N)\right\rVert_\infty$ as \emph{normalized worst objective value} in the remainder of this paper.

\subsection{Neural Architecture Search}\label{sub:nas}
One crucial aspect for the success of deep learning in recent years was the design of novel neural network architectures~\citep{He16,Huan17,Sand18,Szeg16}. However, manually designing such architectures is a cumbersome trial-and-error process. To overcome the need for architectural engineering, \emph{\acl{NAS} (\acs{NAS})} - the process of automatically designing neural network architectures - has arisen as a subfield of automated machine learning~\citep{automl_book}. By now, architectures found by \ac{NAS} have outperformed human-designed architectures on a variety of tasks such as image recognition~\citep{real-arXiv18a}, object detection~\citep{zoph-arXiv18} or dense prediction tasks~\citep{liu_autodeeplab, saikia19}.

We briefly summarize related work here and refer to the survey by Elsken \etal \citep{elsken_survey} for a more thorough literature overview.
Reinforcement learning techniques~\citep{Baker16,Zoph16,Zhong17,zoph-arXiv18} or evolutionary methods~\citep{stanley_evolving_2002, Miikkulainen17, Real17, real-arXiv18a} were employed to search for well performing architectures. As early work required vast amount of computational resources, often in the range of hundreds or even thousands of GPU days~\citep{Zoph16, zoph-arXiv18,  real-arXiv18a}, making \ac{NAS} more efficient was the focus of many researchers, \eg by employing network morphisms \citep{cai-aaai18, cai_path-level_2018, Elsken17}, by sharing weights~\citep{SaxenaV16, bender_icml:2018, Pham18} or by performance prediction~\citep{baker_accelerating_2017, klein-iclr17}. A recent series of work~\citep{liu_darts, xie2018snas, cai2018proxylessnas, zela2019understanding} employed a real-valued relaxation of the discrete architecture search space, enabling gradient-based optimization.  

While the previously discussed approaches solely optimize for a single objective, namely minimizing some error rate, there has also been some work on \emph{multi-objective} neural architecture search \citep{kim-automl17, dong_dpp, mnasnet_tan,lu2019nsganet, Wu2018FBNetHE, Cheng_pareto, hsu_monas, Elsken19}, optimizing other objectives such as network size, latency or energy consumption concurrently. \cite{dong_dpp} extend \citep{liu_progressive_2017} by considering multiple objectives during the model selection step. \cite{lu2019nsganet} employ NSGA-II~\citep{nsga_ii}, a well known multi-objective optimization algorithm, in the context of \ac{NAS}. Instead of actually solving the multi-objective problem, many researchers use scalarization methods, such as the weighted product or sum method \citep{Deb_mo_opt}, to obtain a single objective. This is then optimized via, \eg reinforcement learning~\citep{mnasnet_tan, hsu_monas} or differentiable \ac{NAS}~\citep{Wu2018FBNetHE}. \cite{tea-dnn} use multi-objective Bayesian optimization to search for convolutional cells \citep{zoph-arXiv18}. In this work, we will build up on the multi-objective evolutionary method \pnash~\citep{Elsken19} that exploits cheap-to-evaluate objectives to make the search more efficient. This perfectly fits our application as we will see later as our objectives are solely based on the neural network architecture (and not, e.g., on expensive simulations or trained neural network weights) and thus cheap to compute. We discuss \pnash more detailed in Section~\ref{sec:lemonade}.

\section{Hardware-focused neural architecture design}
\label{sec:hw_focused_nas}
In this section, we introduce our framework for the automated design of error-resilient and hardware-efficient \ac{DNN} architectures. In a first step, we identify optimization goals that typically appear in embedded \ac{DNN} hardware applications and derive corresponding objective functions (Section~\ref{sec:hw_objectives}). In the further course of this paper, these functions serve as input to a multi-objective neural architecture search algorithm (Section~\ref{sec:mo-nas}). Fixed-point quantization is applied as post-processing step after \ac{NAS} to enable efficient \ac{DNN} execution on dedicated hardware accelerators (Section~\ref{sec:quant_methods}).

\subsection{Hardware-specific objectives}
\label{sec:hw_objectives}
We consider four different objectives that are commonly desirable in embedded \ac{DNN} hardware applications, namely high error resilience, low latency, high energy efficiency and a low bandwidth requirement. 

\subsubsection{Error resilience}
In the context of this paper, error resilience is regarded as robustness of the neural network classifier against perturbations in its neuron activation values. Such perturbations can be the result of random hardware faults, such as radiation-induced bit-flips. We measure the degree of perturbation using \ac{BER}, \ie the fraction of flipped bits across all activations of the \ac{DNN}. We define \emph{architecture sensitivity} at a given \ac{BER} as probability for the predicted class output to differ, with and without bit errors. In order to maximize error resilience, we want to minimize architecture sensitivity.

Following the approach in \cite{Scho18} and \cite{Scho19}, we derive an architecture-dependent error sensitivity metric using \ac{LRP}. A key prerequisite in the mathematical framework of \ac{LRP} is the relevance conservation principle \cite{Mont18}. It ensures that the total amount of neuron relevance, which is propagated backwards through the network after the forward pass of inference on an input sample, is conserved in each layer. Consequently, for a group of neurons $k$ and its inputs $j$,
\begin{equation}
\sum_{j}r_j = \sum_{j}\sum_{k}r_{j\leftarrow k} = \sum_{k}\sum_{j}r_{j\leftarrow k} = \sum_{k}r_{k},
\end{equation} 
where $r_j$ and $r_k$ are the relevance values attributed to neurons $j$ and $k$, respectively, and $r_{j\leftarrow k}$ is the amount of relevance propagated backwards from neuron $k$ to neuron $j$. The conservation principle is motivated by the fact that an output activation of neuron $k$ can be completely decomposed into contributions of its input neurons $j$.

The relevance distribution among the neurons in each layer depends on their activations and the synaptic weights \cite{Mont18}. For the initial backpropagation step, the final output neuron relevance of the \ac{DNN} is predetermined by the one-hot encoded target vector belonging to each input sample. This ensures that $\sum_{k}r_{k} = 1$ in each layer. Consequently, for a uniformly randomly drawn neuron in a layer $l$, the expected relevance is
\begin{equation}
\mathrm{E}\left[r_{k}^{(l)}\right] = \frac{1}{n_\mathrm{outputs}^{(l)}},\quad k\sim\mathrm{unif}\{0,n_\mathrm{outputs}^{(l)}-1\},
\end{equation}
where $n_\mathrm{outputs}^{(l)}$ is the total number of neurons in that layer. The observation that a higher average relevance corresponds to a more likely change of the \ac{DNN} classification output suggests that layers with few neurons are more sensitive to errors \cite{Scho18,Scho19}.

\paragraph{Effect of max-pooling.}
Max-pooling is commonly used in some layers of a \ac{DNN}, in order to reduce the output dimensions of that layer \cite{LeCu15}. A max-pooling stage divides the outputs of a layer into subsets and selects the maximum output value out of each subset. We do not regard max-pooling as a separate layer, but consider it as attachment to a layer. If a layer $l$ has max-pooling, the reduced number of output values after the pooling stage is taken to calculate $n_\mathrm{outputs}^{(l)}$.

Additionally, we observed an increased error sensitivity of neurons in layer $l$ if max-pooling is present in the subsequent layer $l+1$. We suppose that this is because information about the input sample is reduced by the pooling stage, but critical errors, which are mostly changes from a low to a high activation value \cite{Li17}, are likely to propagate through. Thus, we obtain an effective error sensitivity of neurons in layer $l$ by multiplication with the pooling factor of layer $l+1$. The pooling factor is the fraction of input to output dimension of the pooling stage and equals $4$ for the max-pooling layers that we use throughout our experiments.  

\paragraph{Effect of merge layers.}
Skip connections, \ie concurrent paths through the network, can improve the training of deep architectures and thus have become popular in state-of-the-art \acp{DNN} \cite{Gu18}. At some point in the network, the parallel paths have to be merged again, which can be done by componentwise addition \cite{He16} or by feature concatenation \cite{Szeg16}. While a concatenation does not affect error propagation, an add layer increases error sensitivity of the \ac{DNN}. There are two reasons for this. Firstly, an add layer involves additional (error prone) load, accumulate and store operations, while concatenation only involves the change of the address range from which data is loaded in the subsequent layers. 

Secondly, the fraction of neurons affected by errors is likely to increase through the add operation. If two inputs with an equal and small fraction of erroneous neurons are added, the resulting fraction of erroneous neurons doubles as long as the error locations in the inputs do not coincide. This can be regarded as doubling the effective error sensitivity of the neurons preceding the add operation.

\paragraph{Architecture sensitivity index.}
The aforementioned insights are now used to define a metric that estimates the error sensitivity of a neural network $N$ solely based on the topology of its architecture. We call this metric \ac{ASI}. It is defined as sum of the expected error sensitivities over all layers $L_N$ of $N$,
\begin{equation}
f_\mathrm{ASI}(N) = \sum_{l \in L_N} \frac{\lambda^{(l)} \zeta^{(l)}}{n_\mathrm{outputs}^{(l)}},
\end{equation}
where $\lambda^{(l)}$ is the max-pooling factor of the succeeding layer $l+1$ (\ie 1 for no pooling) and $\zeta^{(l)}$ is $2$, if $l$ is connected to an add layer, else $1$. 

Concatenations are not counted as extra layers in this sum, while add layers are. Furthermore, supportive functionalities, such as activation function, pooling and batch normalization \cite{Ioff15} are not considered as separate layers, but included in the neuron layers.

We want to emphasize that $f_\mathrm{ASI}$ can be computed very easily, since it only depends on the network topology and does not require any training or other expensive computations. 

\subsubsection{Latency}
\label{sec:latency}
Aside from error resilience, real-time inference with low latency is an additional necessity in many applications. \Acp{AV}, for instance, should be able to derive driving actions from sensory input in less than $\SI{100}{\milli\second}$, in order to surpass human-level perception performance and provide a sufficient level of safety \cite{Lin18}. While low latency can be achieved by employing a parallelized hardware architecture and a high operating frequency, the performance of a \ac{DNN} accelerator is constrained by manufacturing, power consumption, reliability, and flexibility requirements. Thus, a reduction of computational complexity at the algorithm level is desirable.

The roofline model \cite{Will09} is commonly used to describe the attainable computational performance of a \ac{DNN} accelerator \cite{Zhan18c}. It defines two operational domains, which are entered depending on the computational workload of the accelerator. In the memory-bound domain, latency is determined by the amount of data transfer to memory and the available memory bandwidth. In the compute-bound domain, latency can be regarded as being proportional to the number of operations required by the algorithm.

Being compute-bound is preferable over memory-bound operation, since it allows maximum utilization of the available computational resources and highest throughput. Thus, we assume an accelerator, whose memory bandwidth is sufficiently large so that it will predominantly operate in the compute-bound domain for the workloads considered in this paper. We can therefore take the number of operations of the \ac{DNN} as approximate determinant of latency. Furthermore, we regard the number of operations as being solely dependent on the neural network architecture, \ie we do not consider any data-dependent operation reductions.

Our objective function for latency reduction is given by
\begin{equation}
	f_\mathrm{latency}(N) = \sum_{l \in L_N} n_\mathrm{op}^{(l)},
\end{equation}
where $n_\mathrm{op}^{(l)}$ counts the number of operations of layer $l$.

\subsubsection{Energy efficiency}
A further frequent demand on embedded \ac{DNN} accelerators is a low energy consumption per classification inference. This can have mainly two reasons. Firstly, mobile devices have a limited amount of energy storage capacity and thus energy-efficient \ac{DNN} accelerators are required, for example to extend the battery life and range of \acp{AV}. Secondly, embedded devices often have a strict size limitation, which makes it difficult to realize the necessary heat dissipation. As the thermal leakage power of an accelerator directly depends on the number of classifications per second and the energy per classification, energy efficiency is desirable to enable high classification throughput.

Energy consumption of \ac{DNN} accelerators is dominated by data transfers to and from memory \cite{Sze17}. This is due to the large amount of parameters and intermediate data outputs of typical large-scale \acp{DNN}. 

According to Horowitz \cite{Horo14}, energy consumption for off-chip dynamic \ac{RAM} access is about two orders of magnitude higher than for internal cache accesses and arithmetic operations. While some hardware designers increase energy efficiency by integrating huge on-chip static \acp{RAM} in their \ac{DNN} accelerator (\eg \cite{TESLA_FSD}), this approach is not feasible in every case. In this paper, we assume an accelerator with small on-chip buffer (such as \cite{Chen17}), so that a layerwise data transfer to and from off-chip memory is necessary, which dominates energy consumption. 

Consequently, to maximize energy efficiency, our objective is to minimize data transfer to and from memory per inference. We neglect the number of operations in this calculation because of its limited influence on energy consumption and since it is already part of the latency minimization objective function. To determine the data transfer of a layer, we assume that each input and weight parameter of the layer is loaded once from external memory and each output is written back once. Furthermore, we assume that the same bit-width is used to represent all activations and parameters of the network.

Our objective function for minimizing energy consumption is thus given by the sum of layerwise input, output, and parameter data word transfers over the whole network,
\begin{equation}
	f_\mathrm{energy}(N) = \sum_{l \in L_N} (n_\mathrm{inputs}^{(l)} + n_\mathrm{outputs}^{(l)} + n_\mathrm{params}^{(l)}),
\end{equation}
where $n_\mathrm{inputs}^{(l)}$ and $n_\mathrm{outputs}^{(l)}$ count the number of input neurons and output neurons, respectively, and $n_\mathrm{params}^{(l)}$ counts the number of parameters of layer $l$.

\subsubsection{Bandwidth requirement}
As described in Section~\ref{sec:latency}, we assume the accelerator for which we optimize \ac{DNN} architectures to operate predominantly in the compute-bound domain of the roofline model. In order to guarantee compute-bound operation, the accelerator has to provide a certain maximum bandwidth to memory. It is desirable to keep this bandwidth requirement within bounds to simplify the accelerator architecture.

The required memory bandwidth can vary for the different layers of a \ac{DNN}. We employ the ratio between data transfers and operations of a layer as estimator for its bandwidth requirement. The intuition behind this is that a low number of operations is related to a short processing time of the layer and consequently a high bandwidth is required to be able to perform the necessary data movements in that given time.

We define an overall objective function to optimize neural architectures for a low bandwidth requirement by adding up the data-computation ratios of all layers. Thus our objective function for minimizing the bandwidth requirement is given by the \ac{ADCR}:
\begin{equation}
	f_\mathrm{ADCR}(N) = \sum_{l \in L_N} \frac{n_\mathrm{inputs}^{(l)} + n_\mathrm{outputs}^{(l)} + n_\mathrm{params}^{(l)}}{n_\mathrm{op}^{(l)}}.
\end{equation}

\subsection{Multi-objective NAS}\label{sec:mo-nas}

In the following, we introduce \pnash, a \textbf{L}amarckian \textbf{E}volutionary algorithm for \textbf{M}ulti-\textbf{O}bjective \textbf{N}eural \textbf{A}rchitecture \textbf{DE}sign \citep{Elsken19}, that we will use in our later experiments to automatically design well\hyp{}performing, error-resilient, and hardware-efficient architectures.

\subsubsection{\pnash}
\label{sec:lemonade} 
\pnash maintains a population $\pf$ of neural networks $N$. This population is improved over the course of the algorithm with respect to multi-objective optimization problem 
   $\min_{N \in \NN} \moaf(N),$ 
where $\NN$ denotes a suitable space of neural network architectures (see Section~\ref{sec:searchspace}) and the objective function
\begin{equation}
\moaf(N) = (\fexp(N), \fcheap(N)  )^\top \in \mathds{R}^m \times \mathds{R}^n
\end{equation}
is split into expensive-to-evaluate objectives $\fexp(N) \in \mathds{R}^m$ (in our case: the validation error, only obtainable by expensive training) and cheap-to-evaluate objectives $\fcheap(N) \in \mathds{R}^n$ (in our case: the objectives defined in Section~\ref{sec:hw_objectives}). The population $\pf$ is chosen to comprise all non-dominated networks with respect to $\moaf$, \ie the population approximates the Pareto front. \pnash exploits that $\fcheap$ is cheap to evaluate in order to bias the sampling of children towards areas of the Pareto front of $\fcheap$ that are sparsely populated. While $\fcheap$ is evaluated many times in \pnash, $\fexp$ is evaluated only a few times for promising networks that are likely to improve the approximation of the Pareto front.

In every iteration of \pnash, firstly parent networks are sampled with respect to some probability distribution (discussed later) that is solely based on the cheap objectives. By applying mutations to the parents (such as adding or removing a layer, see Section~\ref{sec:mutations} for a detailed description), children are generated. In a second sampling stage, a subset of all generated children is selected, again solely based on cheap objectives, and solely this subset is evaluated on the expensive objectives $\fexp$. Lastly, \pnash computes the Pareto front from the current generation and the subset of generated children, yielding the next generation. The described procedure is repeated for a pre-specified number of iterations.
  
\paragraph{The sampling distribution.} The sampling distribution is designed to only depend on the cheap objectives and to guide the search towards sparsely crowded regions in the current Pareto front. In order to achieve this, \pnash computes a kernel density estimator $\pkde$ on the cheap objective values $\{ \fcheap(N) | N \in \pf\}$ of the current population. Then, for both sampling stages (\ie (i) the probability for choosing a network $N$ as a parent as well as (ii) the probability of a generated child $N$ being part of the subset), \pnash uses a sampling distribution anti-proportional to $\pkde$:
\begin{equation}\label{eq:parentdist}
    p(N) = \frac{c}{\pkde(\fcheap(N))},
\end{equation}
with a proper normalizing constant $c$. Therefore, networks in sparsely populated regions of the Pareto front are more likely to be chosen as parents and generated children lying in sparsely populated regions of the Pareto front are more likely to be evaluated on $\moaf$. The motivation behind also choosing parents in less crowded regions is that mutations do not change the network drastically, hence children are expected to have similar objective values as their parents. By this sampling distribution and the two-staged sampling strategy, \pnash generates and evaluates more children that are more likely to improve the current approximation of the Pareto front rather then just evaluating the cheap objective $\fexp(N)$ for all children, making it more efficient than off-the-shelf multi-objective optimization algorithms. We highlight that all objectives from Section~\ref{sec:hw_objectives} are cheap-to-evaluate as they all solely depend on the neural network architecture and not, \eg on the weights of the network only obtainable by expensive training. Hence, \pnash is a perfect fit for our purpose. For more details, we refer the reader to the original work~\citep{Elsken19}.

\subsubsection{Search space and mutations within \pnash}
\label{sec:mutations}\label{sec:searchspace}

In this work, we focus on \ac{NAS} for image classification tasks. \Acp{ConvNet} are the predominantly used type of \ac{DNN} in this domain \cite{LeCu15}. However, in the recent years, the number of variations and design choices for \ac{ConvNet} architectures has significantly grown (see \eg \cite{Gu18} for an overview). We limit the search space of \pnash to a number of predefined building blocks, hyperparameters and allowed mutations for two reasons. Firstly, support for a limited set of building blocks requires less flexibility of the underlying hardware. This enables the use of more efficient dedicated \ac{DNN} accelerators instead of general purpose hardware. Secondly, the space of feasible architectures $\NN$ rapidly grows with each additional variation that is allowed. This combinatorial explosion slows down the convergence of \ac{NAS}, which is why a reasonable limitation of the search space has to be chosen.

We now describe the set of mutations that are used by \pnash in our experiments to generate child networks.
	\begin{enumerate}[leftmargin=*]
	\item Insert a convolutional layer with batch normalization \cite{Ioff15} and \ac{ReLU} activation \cite{Glor11}. The layer is inserted at a random position and its number of filters is chosen to match the number of filters of the preceding layer. The kernel height $h$ and width $w$ of the convolutional filter are randomly sampled: $(h,w) \in \{(3,3), (5,5), (7,7), (9,9)\}$.
	\item Increase the number of filters of a randomly chosen convolution by a randomly chosen factor $\in \{2, 4\}$. A maximum of $1100$ filters is allowed.
	\item Add a skip connection. We allow skip connection either by concatenation \cite{Szeg16} or by addition \cite{He16}.
	\item Remove a randomly chosen layer or a skip connection.
	\item Prune a randomly chosen convolutional layer (\ie remove $1/2$ or $1/4$ of its filters). A minimum of $15$ filters is allowed.
	\item Replace a randomly chosen convolution by a depthwise separable convolution \cite{Chol17}.
	\end{enumerate}
Note that by random we always mean uniformly at random. We highlight that the first three operations in general increase objectives such as network's size or energy consumption, but likely also decrease objectives such as the error, while the last three operations in general decrease the firstly mentioned objectives, but increase the lastly objectives. Consequently, these mutations are suitable for multiple, opposing objectives.

To further speed up \ac{NAS}, the authors of \pnash propose to apply these mutations as \emph{network morphisms} \citep{DBLP:journals/corr/ChenGS15, pmlr-v48-wei16}. Network morphisms are function-preserving operators on neural networks, \ie a network morphism maps a neural network $N^w$ with weights $w$ to another neural network $\tilde{N}^{\tilde{w}}$ with weights $\tilde{w}$ so that for every input $x$ to the network $N^w(x) = \tilde{N}^{\tilde{w}}(x)$. Effectively this means that, when utilizing network morphisms as mutations to generate children, children \emph{do not need to be trained from scratch} but rather just fine-tuned as children \emph{by design} have the same error as their parent. This can be interpreted as Lamarckian inheritance in the context of evolutionary algorithms, where Lamarckism refers to a mechanism which allows passing skills acquired during an individual's lifetime (\eg by means of learning), on to children by means of inheritance. The equality $N^w(x) = \tilde{N}^{\tilde{w}}(x)$ can be achieved by properly choosing $\tilde{w}$. For example, if one wants to insert a linear layer at an arbitrary position in a network, equality can be achieved by simply initializing the linear layer as an identity mapping. Mutations 1--3 from above can all be formulated as a network morphism (see \cite{Elsken19} for details). Mutations 4--6, on the other hand, cannot be framed as network morphisms, as they all generally decrease the network's capacity and equality cannot be guaranteed. Instead, Elsken \etal \citep{Elsken19} propose \emph{approximate} network morphisms to find proper initialization for these cases. Approximate network morphisms essentially copy the weights of layers not affected by structural changes and train affected layers via knowledge distillation~\citep{distillation}. 

\subsection{Fixed-point quantization}
\label{sec:quant_methods}
Neural network training algorithms usually rely on data representations and computations with high numerical precision, for example a 32-bit floating-point format, typically used in \acp{GPU}. However, after training, a reduced-precision number format can be used for inference on a dedicated \ac{DNN} accelerator to reduce energy consumption and bandwidth \cite{Lin16b}. In this context, an 8-bit fixed-point format is a common choice in embedded and mobile devices \cite{Jaco18}. Hence, to deploy a \ac{DNN} on an embedded device after training on a \ac{GPU}, weights, biases and activations need to be transformed from a floating-point to a fixed-point number format. This procedure is denoted by network \emph{quantization}. We apply network quantization as post-processing step after \acl{NAS} with \pnash.

To quantize a real value $\chi$ to a signed fixed-point value $\chi_q$ using $B$ bit, we determine
\begin{equation}
\chi_q = \text{clip}\left( \text{round}\left( \frac{\chi}{\Delta} \right), -2^{B-1}, 2^{B-1} \right) \Delta,
\end{equation}
where $\Delta$ denotes the step size, i.\,e.\ the smallest distance between two quantization sampling points of $\chi$. In other words, $\Delta$ corresponds to the value of the \ac{LSB}.

In \cite{Vanh11}, a simple method to find a suitable step size for a given data distribution in \acp{DNN} with sigmoid activations is introduced. It determines the step size $\Delta$ based on the maximum range of a distribution according to
\begin{equation}\label{eq:quant_max_range}
\Delta = \frac{\text{max}\left( \vert \chi \vert \right)}{2^{N-1}-1}.
\end{equation}
In the following, we refer to this quantization method as \emph{MaxRange}. However, modern \acp{DNN} commonly use unbounded activation functions, such as \ac{ReLU}, and thus may entail data distributions with far outliers. Since the quantization range is adapted to the maximum value, the step size $\Delta$ is maximal and consequently leads to a coarse sampling of smaller values. Moreover, as data distributions in \acp{DNN} typically follow a Gaussian distribution, \eqref{eq:quant_max_range} leads to a coarse sampling of a large number of values.

A quantization method which specifically targets this problem has been introduced in \cite{Voge19b}. Here, parameters and activations are quantized by minimizing the effect of the quantization error $\delta=\chi-\chi_q$ in the network. In a neural network, the output value $y$ of a neuron with a rectifying unit $\Phi(\cdot)$, bias $b$, weights $w$ and input values $x$ is determined by
\begin{equation}\label{eq:neuron_output}
y=\Phi\left( b+\sum wx \right).
\end{equation}
For the purpose of measuring the influence of the quantization error of inputs ($\delta_x$), weights ($\delta_w$) and biases ($\delta_b$), we define $\tilde{y}$ as the resulting neuron output when quantities of \eqref{eq:neuron_output} are quantized. More precisely, $\tilde{y}_w$ is defined as the neuron output determined with quantized weights $w_q$ where activations and biases remain in a 32\,bit floating-point number format. $\tilde{y}_x$ and $\tilde{y}_b$ are defined accordingly. The step sizes $\Delta^{(l)}$ are then individually determined for each layer by
\begin{equation}\label{eq:step_size_propQE}
\begin{aligned}
\Delta_w^{(l)} &= \underset{\Delta^{(l)}_w}{\text{arg\,min}} \left\lvert y^{(l)}-\tilde{y}^{(l)}_w \right\rvert^2,\\
\Delta_x^{(l)} &= \underset{\Delta^{(l)}_x}{\text{arg\,min}} \left\lvert y^{(l)}-\tilde{y}^{(l)}_x \right\rvert^2~~\text{and}\\
\Delta_b^{(l)} &= \underset{\Delta^{(l)}_b}{\text{arg\,min}} \left\lvert y^{(l)}-\tilde{y}^{(l)}_b \right\rvert^2.\\
\end{aligned}
\end{equation}
We additionally constrain the step sizes to a power-of-two value, i.\,e.\ $\Delta\in\{2^z\,\vert\,z\in\mathbb{Z}\}$, to enable a direct fixed-point operation in a hardware accelerator. In the rest of the paper, this quantization method is referred to as \emph{minimal propagated quantization error (MinPQE)}.

\section{Experiments}
\label{sec:experiments}
\subsection{Experimental setup}
To evaluate our methods, we use two common image classification benchmarks. Firstly, CIFAR-10 \cite{Kriz09} is used, which consists of $32\times 32$ pixel RGB images divided into ten distinct classes. The samples are divided into 50\,000 training and 10\,000 test samples. Out of the training set, 5000 samples are used for validation during neural architecture search.

Secondly, \ac{GTSRB} \cite{Stal12} is used, which contains RGB images of 43 different types of traffic signs. The images of this benchmark are scaled to a resolution of $48\times 48$ pixels before they are fed into the classifier. The dataset has 39\,210 training samples, out of which 4010 are separated for classification validation. An additional set of 12\,630 images is used for measuring final test error rates.

Unless otherwise noted, we use the same hyperparameter setup for both benchmarks. We run \pnash for 300 evolutionary iterations. The algorithm is initialized with a population of 15 manually chosen trivial network architectures with different numbers of convolutional layers and kernel shapes. For \ac{DNN} training, we use \ac{SGD} with cosine annealing \cite{Losh17}, momentum of $0.9$ and a weight decay of $0.0005$. The learning rate for each training phase during architecture search is initialized with $0.01$. The training batch size is set to 64 throughout our experiments. Furthermore, we apply commonly used data augmentations during training \cite{Losh17}. However, we leave out horizontal image flips for \ac{GTSRB}, since they would change the meaning of some traffic signs. In addition, we use mixup \cite{Zhan18d} and cutout \cite{DeVr17} for further training data augmentation.

The final population sizes of the CIFAR-10 and \ac{GTSRB} models are 439 and 238, respectively. From each of these, the 50 architectures with best validation error rates are selected and each of these is trained from scratch on the set of training and validation images for 200 epochs. The learning rate is initialized with $0.025$ in this case and all other hyperparameters stay the same. Classification error is evaluated on the separate test set after the training. Subsequently, we quantize the networks' weights and activations to an 8-bit fixed-point representation using the MaxRange and MinPQE methods described in Section~\ref{sec:quant_methods} for further evaluations.

\subsection{Error simulations}
\label{sec:fi_framework}
\begin{figure}[!ht]
	\centerline{\includegraphics[width=0.48\textwidth]{{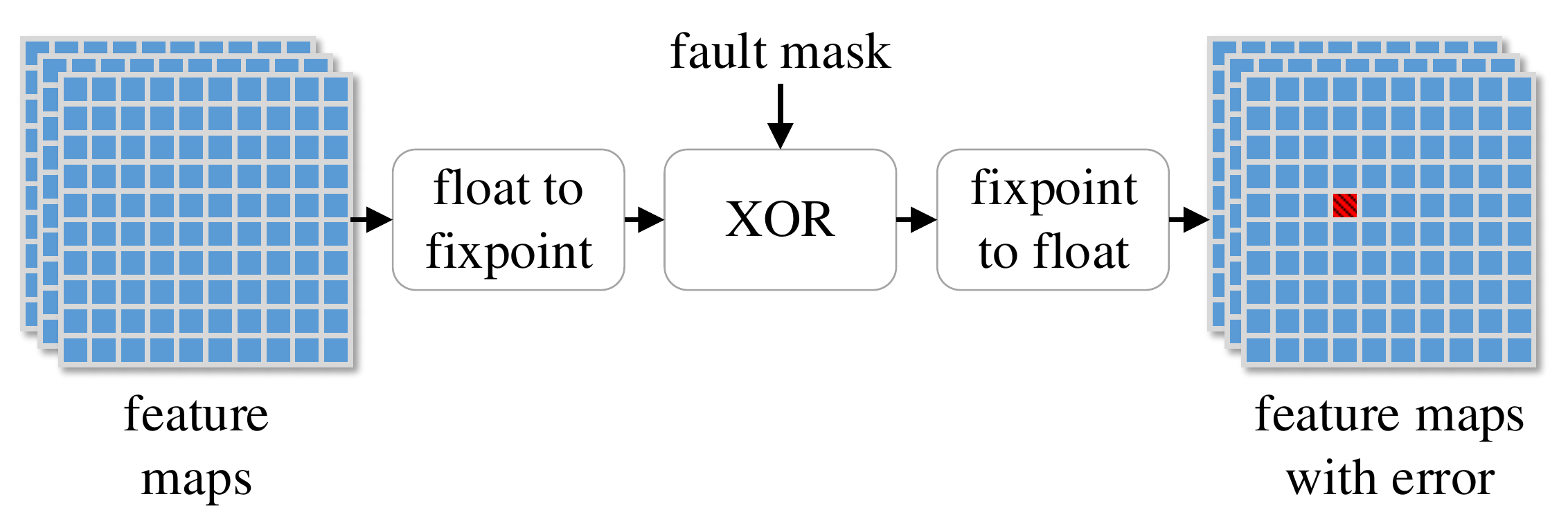}}}
	\caption{Steps performed by fault injection framework between the computation of two neural network layers \cite{Scho19}.}
	\label{fig:fault_injection}
\end{figure}

Random bit-flip error simulations are used to evaluate the actual resilience of the obtained set of neural networks. For this purpose, we use the fault simulation framework that has been previously described in \cite{Scho19}. The framework builds up on the Keras \cite{Chol15} \ac{DNN} library with TensorFlow back-end \cite{Abad15}. This allows for performing fast bit-level fault injections in the neuron activation outputs (feature maps) of a \ac{ConvNet}. Most of the computation workload required for the simulation can be efficiently computed on a \ac{GPU}. The framework automatically adds some operations behind each neuron output stage of a given \ac{ConvNet}, which emulate a fixed-point format and allow for a bit-wise fault injection in the neuron output memory by applying a definable Boolean fault mask (see \figurename~\ref{fig:fault_injection}).

\subsection{Results}

\subsubsection{Trade-off analysis between objectives}

\begin{table*}[htbp]
	\caption{Properties of models that optimize a respective cost dimension. E.g.\ model BestASI (second row) denotes the optimizer of the ASI objective. Bold numbers indicate minimal values among the selected 50 models with best validation error. Data transfer and accumulated data-computation ratio are calculated taking 8-bit fixed-point quantization of activations and weights into account.}
	\begin{center}
		\begin{tabularx}{\textwidth}{lXXXXXcXXXXX}
			\toprule
			\textbf{Model}&\multicolumn{5}{c}{\textbf{Optimized Quantities}}&&\multicolumn{5}{c}{\textbf{Other Quantities}}\\
			\cmidrule{2-6}
			\cmidrule{8-12}
			&\rotatebox{90}{\parbox{2.2cm}{Architecture \\Sensitivity \\Index ($\times 10^{-3}$)}}  
			&\rotatebox{90}{\parbox{2.2cm}{Validation Set \\Error Rate (\%)}}
			&\rotatebox{90}{\parbox{2.2cm}{Operations \\(GOp/Frame)}}
			&\rotatebox{90}{\parbox{2.2cm}{Data Transfer \\(MB/Frame)}}  
			&\rotatebox{90}{\parbox{2.2cm}{Acc. Data- \\Computation \\Ratio (B/Op)}}
			&
			&\rotatebox{90}{\parbox{2.2cm}{Normalized \\Worst Objec- \\tive Value}} 
			&\rotatebox{90}{\parbox{2.2cm}{Number of \\Parameters \\($\times 10^{6}$)}} 
			&\rotatebox{90}{\parbox{2.2cm}{Test Set Error \\ Rate (\%) \\(32b float)}}
			&\rotatebox{90}{\parbox{2.2cm}{Test Set Error \\ Rate (\%) \\(8b MaxRange)}}
			&\rotatebox{90}{\parbox{2.2cm}{Test Set Error \\ Rate (\%) \\(8b MinPQE)}} \\
			\midrule
			\textbf{CIFAR-10} & & & & & & & & & & & \\
			WorstASI & 8.891 & 9.20 & 0.050 & 0.672 & 7.279 && 1.000 & 0.344 & 7.31 & 7.58 & 7.52 \\
			BestASI & \textbf{0.336} & 9.16 & 0.420 & 2.112 & 1.422 && 0.959 & 1.645 & 6.95 & 6.91 & 6.87 \\
			BestValErr & 4.267 & \textbf{6.52} & 0.186 & 2.381 & 10.230 && 0.996 & 1.489 & 5.48 & 5.33 & 5.41 \\
			BestEfficiency & 1.750 & 9.18 & \textbf{0.049} & \textbf{0.665} & 10.264 && 1.000 & \textbf{0.337} & 6.54 & 6.68 & 6.61 \\
			BestADCR & 0.336 & 9.30 & 0.429 & 2.122 & \textbf{1.150} && 0.993 & 1.654 & 6.42 & 6.57 & 6.47 \\
			BalOpt & 0.970 & 7.56 & 0.127 & 1.668 & 4.241 && \textbf{0.371} & 1.330 & 5.72 & 5.66 & 5.63 \\
			\midrule
			\textbf{GTSRB} & & & & & & & & & & & \\
			WorstASI & 8.120 & 0.45 & 0.045 & 0.478 & 10.218 && 1.000 & 0.101 & 2.53 & 2.66 & 2.64 \\
			BestASI & \textbf{0.109} & 0.30 & 0.490 & 1.220 & 1.058 && 0.501 & 0.865 & 2.60 & 2.64 & 2.60 \\
			BestValErr & 0.217 & \textbf{0.00} & 0.966 & 4.629 & 4.081 && 1.000 & 3.200 & 0.90 & 1.08 & 0.99 \\
			BestEfficiency & 0.651 & 0.45 & \textbf{0.012} & \textbf{0.181} & 1.166 && 0.600 & \textbf{0.041} & 1.32 & 1.41 & 1.41 \\
			BestADCR & 0.145 & 0.12 & 0.600 & 3.161 & \textbf{1.048} && 0.670 & 2.833 & 2.50 & 2.61 & 2.62 \\
			BalOpt & 0.326 & 0.20 & 0.126 & 0.676 & 1.057 && \textbf{0.267} & 0.513 & 2.78 & 2.84 & 2.81 \\
			\bottomrule
		\end{tabularx}
		\label{tab:moo_results}
	\end{center}
\end{table*}

\tablename{}~\ref{tab:moo_results} lists the properties of certain \ac{DNN} architectures $N$ obtained for both benchmarks, CIFAR-10 and \ac{GTSRB}. The selected models are the ones that minimize each an individual objective function $f_i(N)$ (BestASI, BestValErr, BestEfficiency and BestADCR), the model with maximum error sensitivity (WorstASI) as well as the model with lowest normalized worst objective value (see Section~\ref{sub:moo}) $\left\lVert\bar{\moaf}(N)\right\rVert_\infty$, \ie the balanced optimizer of all objectives (BalOpt). The BestEfficiency models actually minimize both $f_\mathrm{latency}(N)$ (\ie operations) and $f_\mathrm{energy}(N)$ (i.e. data transfer). This indicates a correlation between the two quantities. The respective models are also the smallest in terms of weight parameters.

It can be seen in \tablename{}~\ref{tab:moo_results} that choosing a \ac{DNN} with minimal cost in one objective often leads to the outcome that at least one other objective is close to its worst value. This is especially the case for CIFAR-10, where $\left\lVert\bar{\moaf}(N)\right\rVert_\infty$ is $1$ or close to $1$ for all single-objective optimizers, BestASI, BestValErr, BestEfficiency, and BestADCR. The optimal trade-off models (BalOpt), however, come quite close to the ideal point, with normalized distances of $0.371$ (CIFAR-10) and $0.267$ (\acs{GTSRB}).

Another aspect visible in \tablename{}~\ref{tab:moo_results} is that 8-bit quantization does not significantly increase test set classification error rates of the models in comparison to the 32-bit float case (in some cases the error is even smaller after quantization). The differences between the MaxRange and MinPQE quantization methods with respect to test error rate are marginal. 

\begin{figure*}[!ht]
	\centering
	\subfloat[][]{\includegraphics[width=0.23\textwidth]{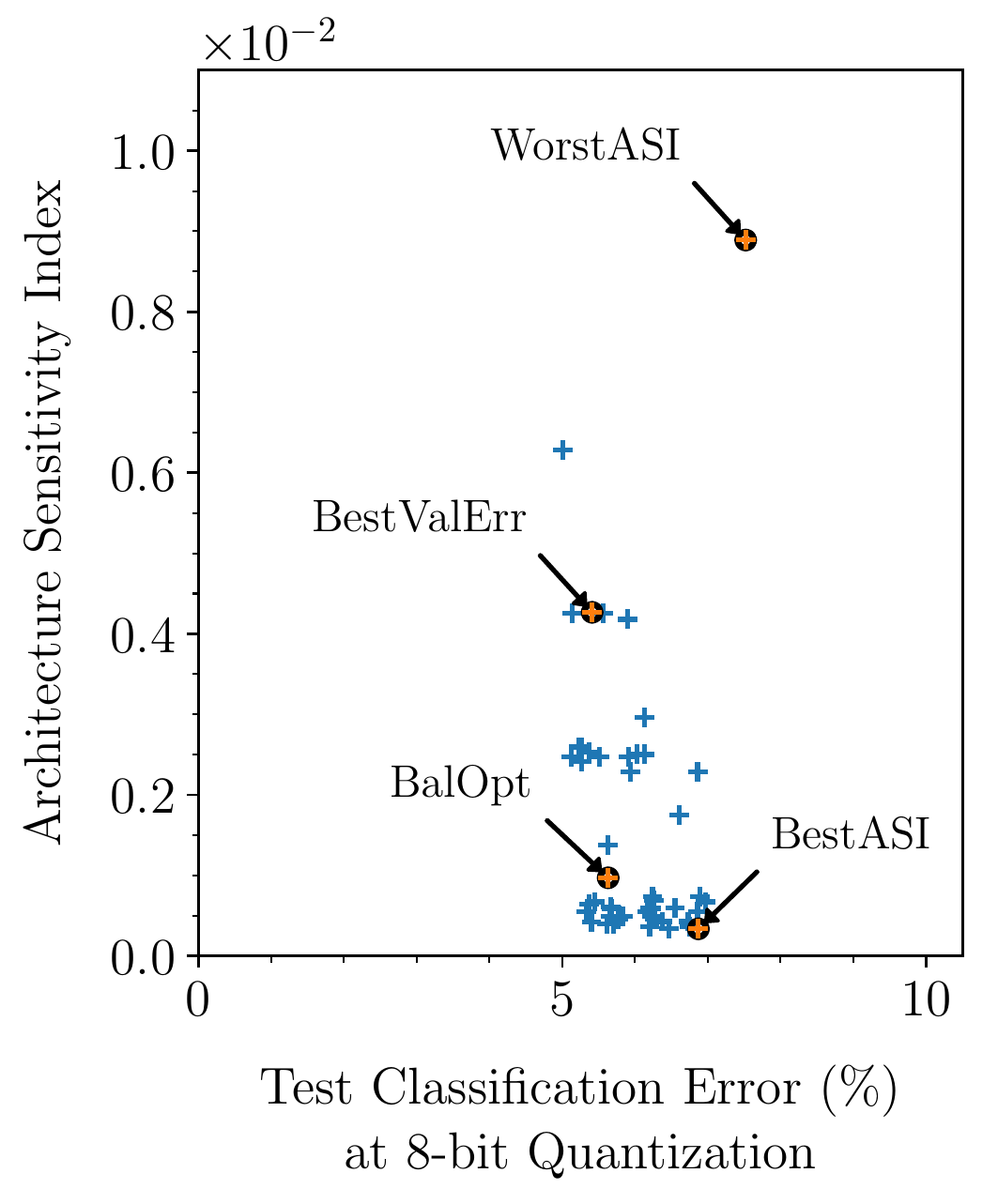}}
	\quad
	\subfloat[][]{\includegraphics[width=0.23\textwidth]{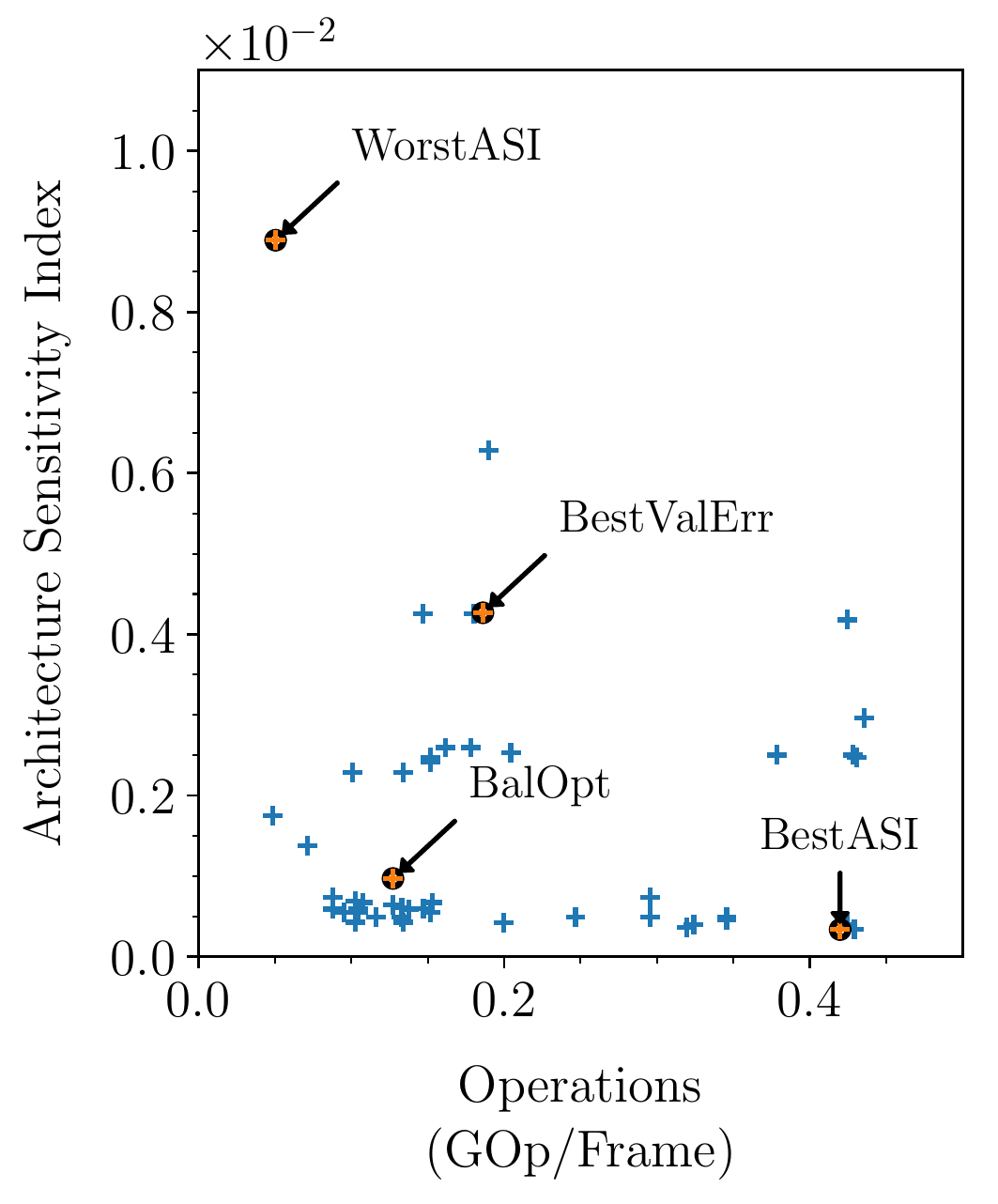}}
	\quad
	\subfloat[][]{\includegraphics[width=0.23\textwidth]{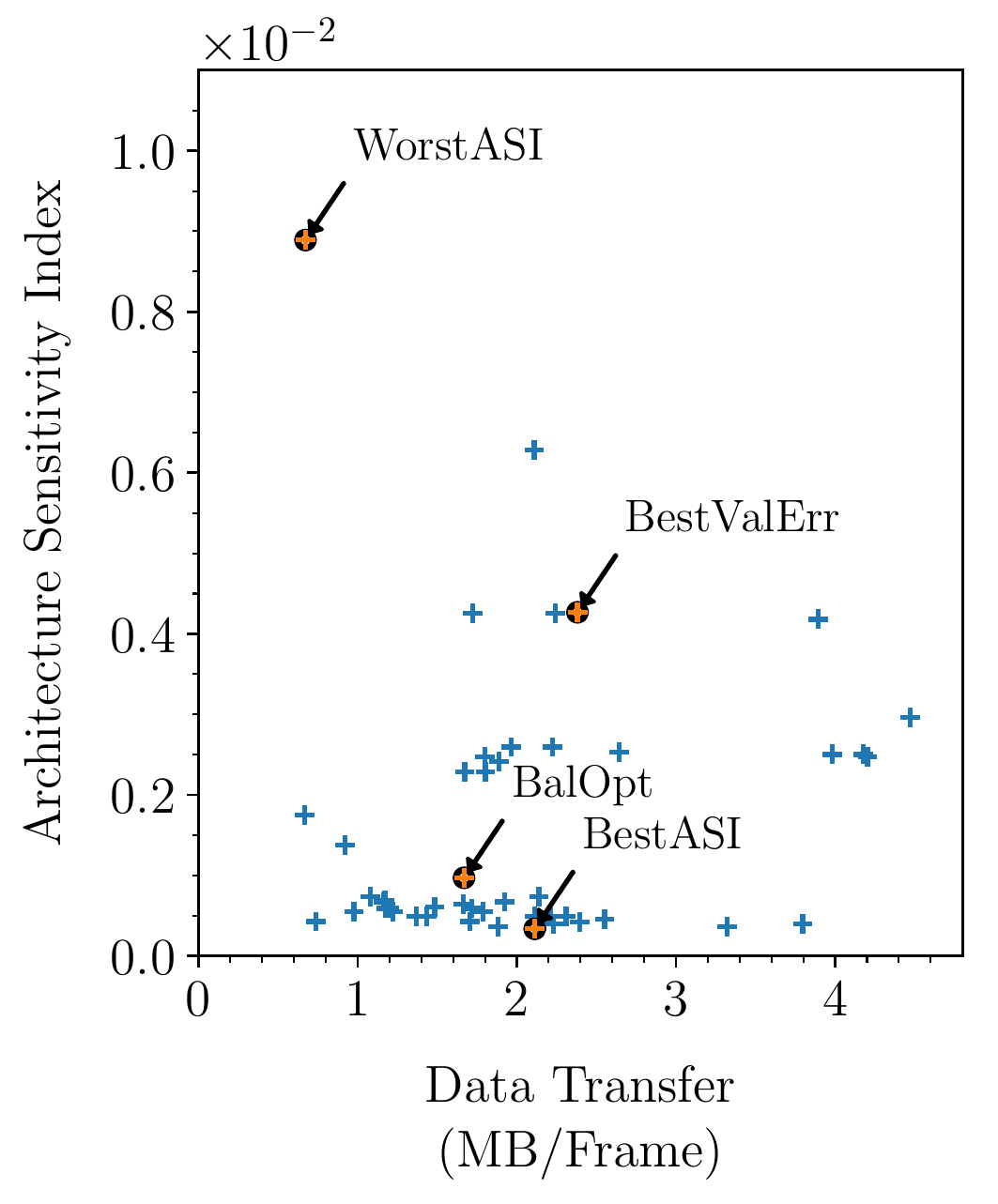}}
	\quad
	\subfloat[][]{\includegraphics[width=0.23\textwidth]{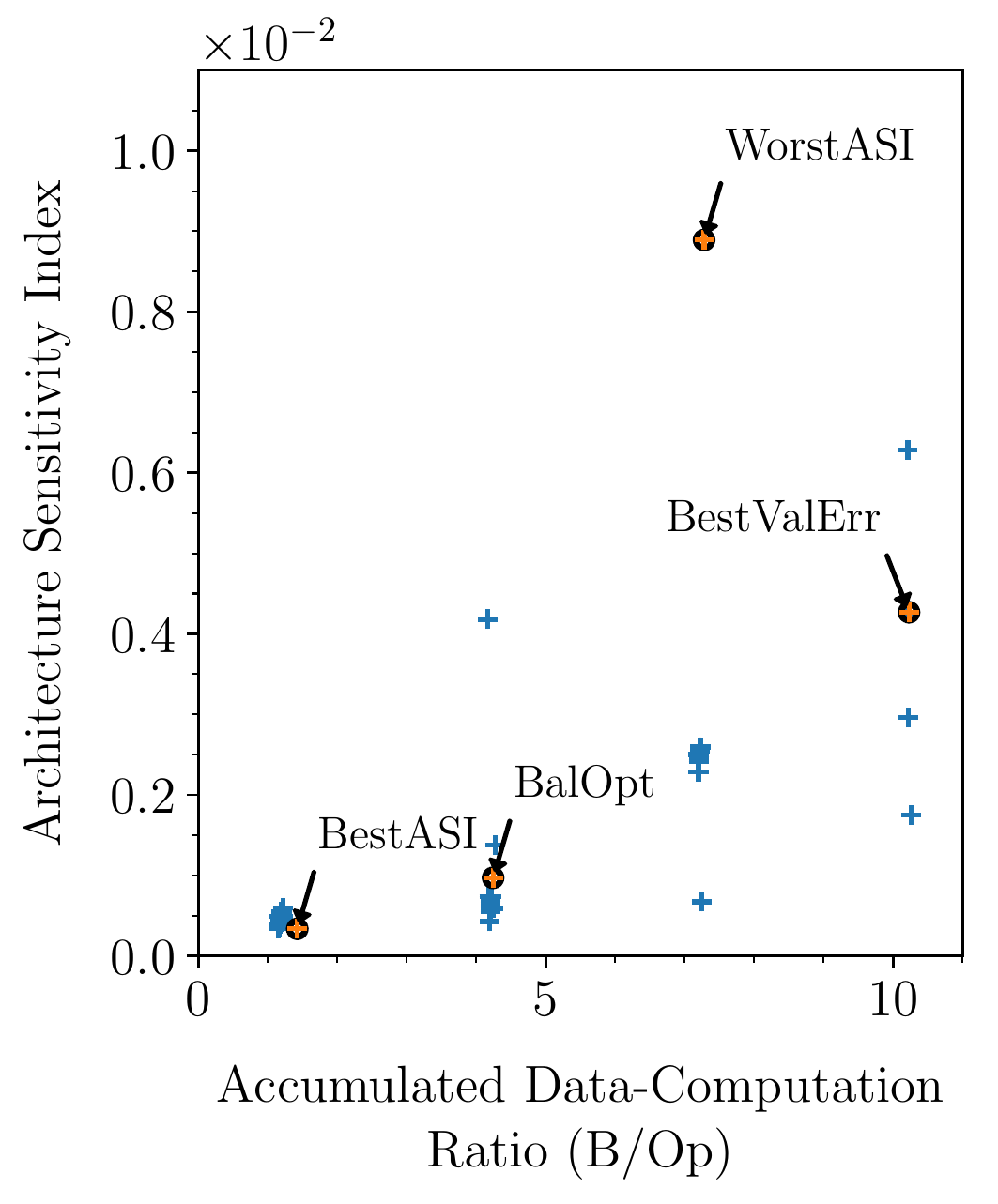}}
	\caption{Pairwise comparison of ASI with each of the other objective function outcomes for 50 Pareto-optimal architectures on CIFAR-10.}
	\label{fig:moo_cifar10_pareto}
\end{figure*}

\begin{figure*}[!ht]
	\centering
	\subfloat[][]{\includegraphics[width=0.23\textwidth]{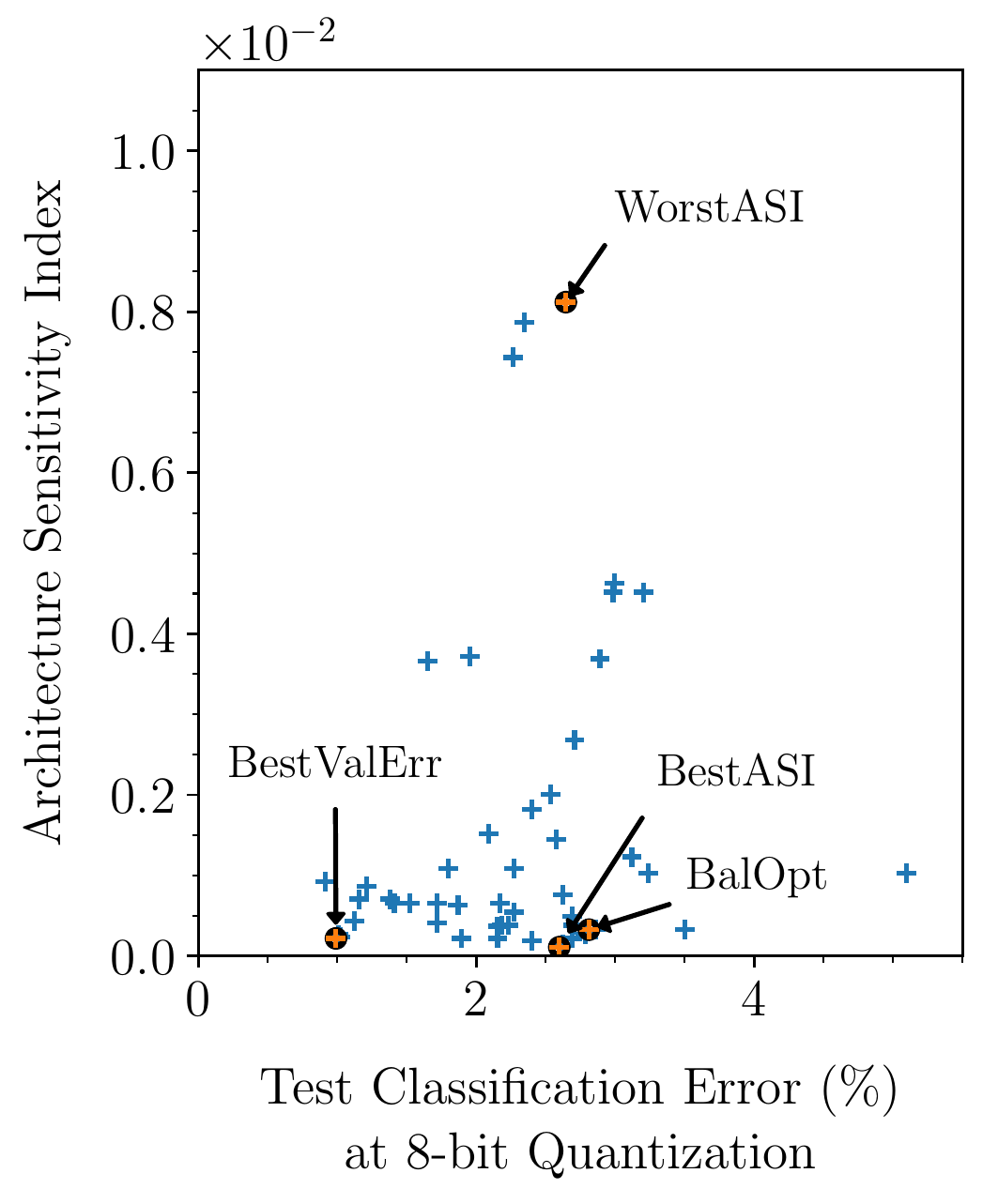}}
	\quad
	\subfloat[][]{\includegraphics[width=0.23\textwidth]{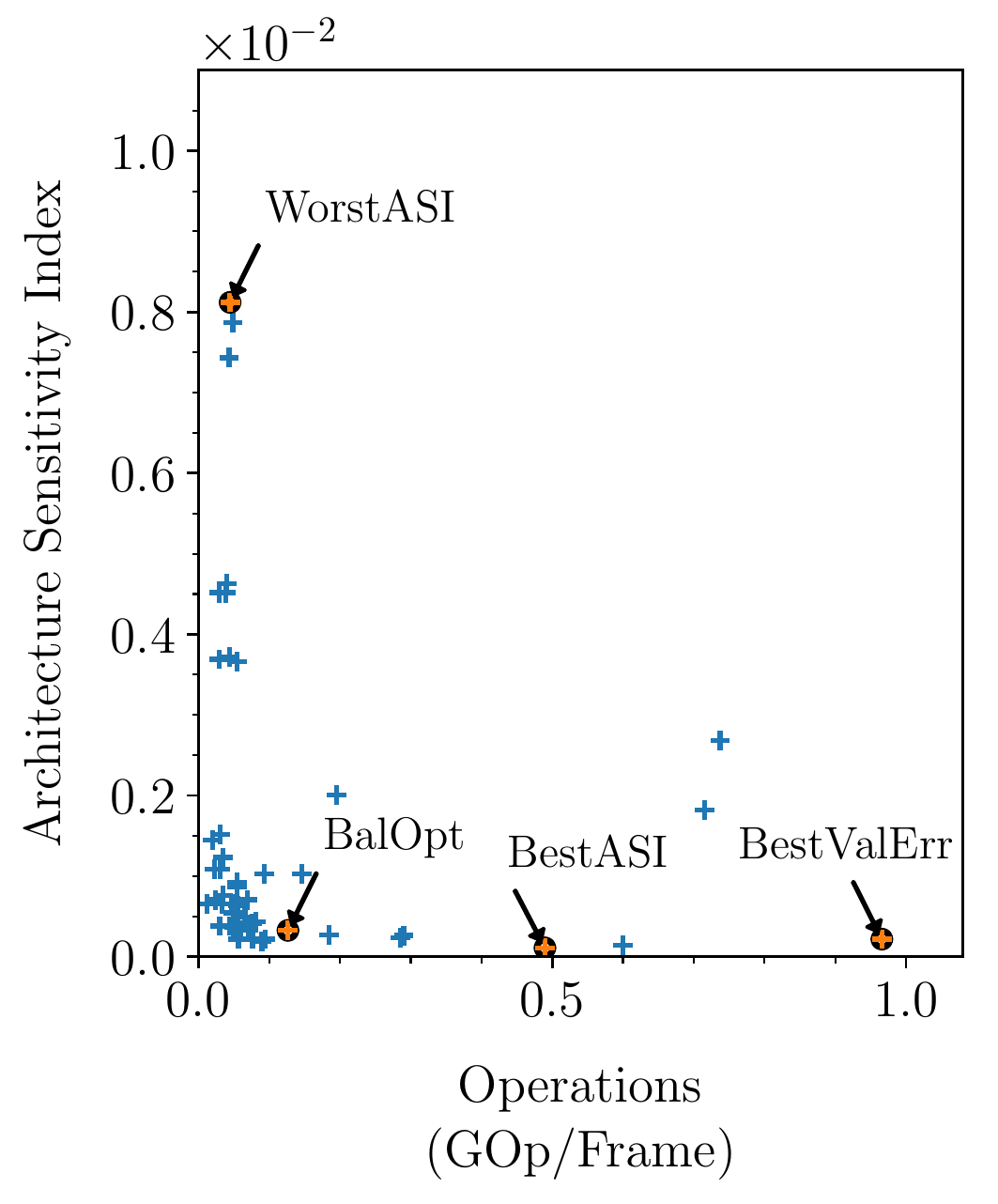}}
	\quad
	\subfloat[][]{\includegraphics[width=0.23\textwidth]{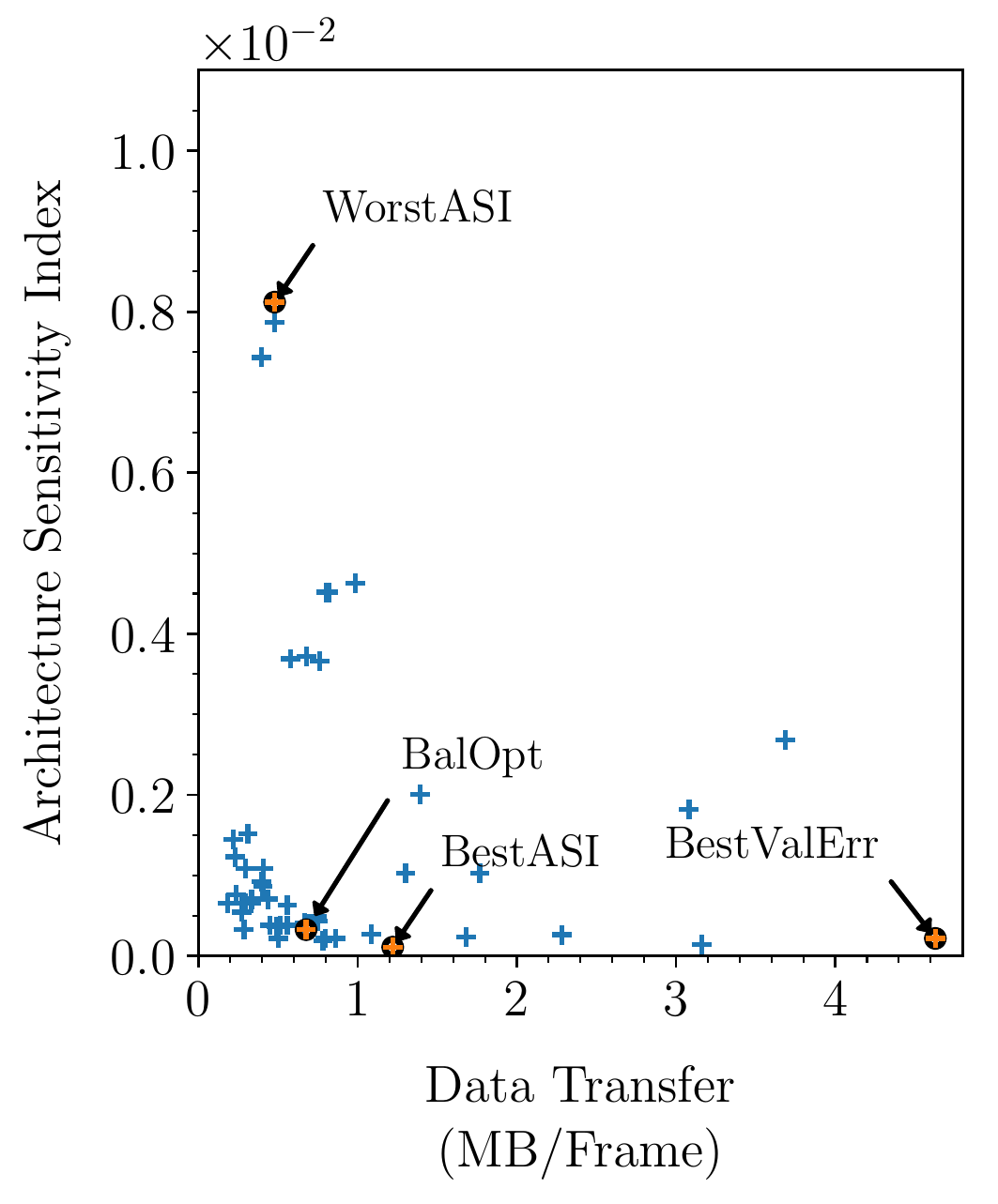}}
	\quad
	\subfloat[][]{\includegraphics[width=0.23\textwidth]{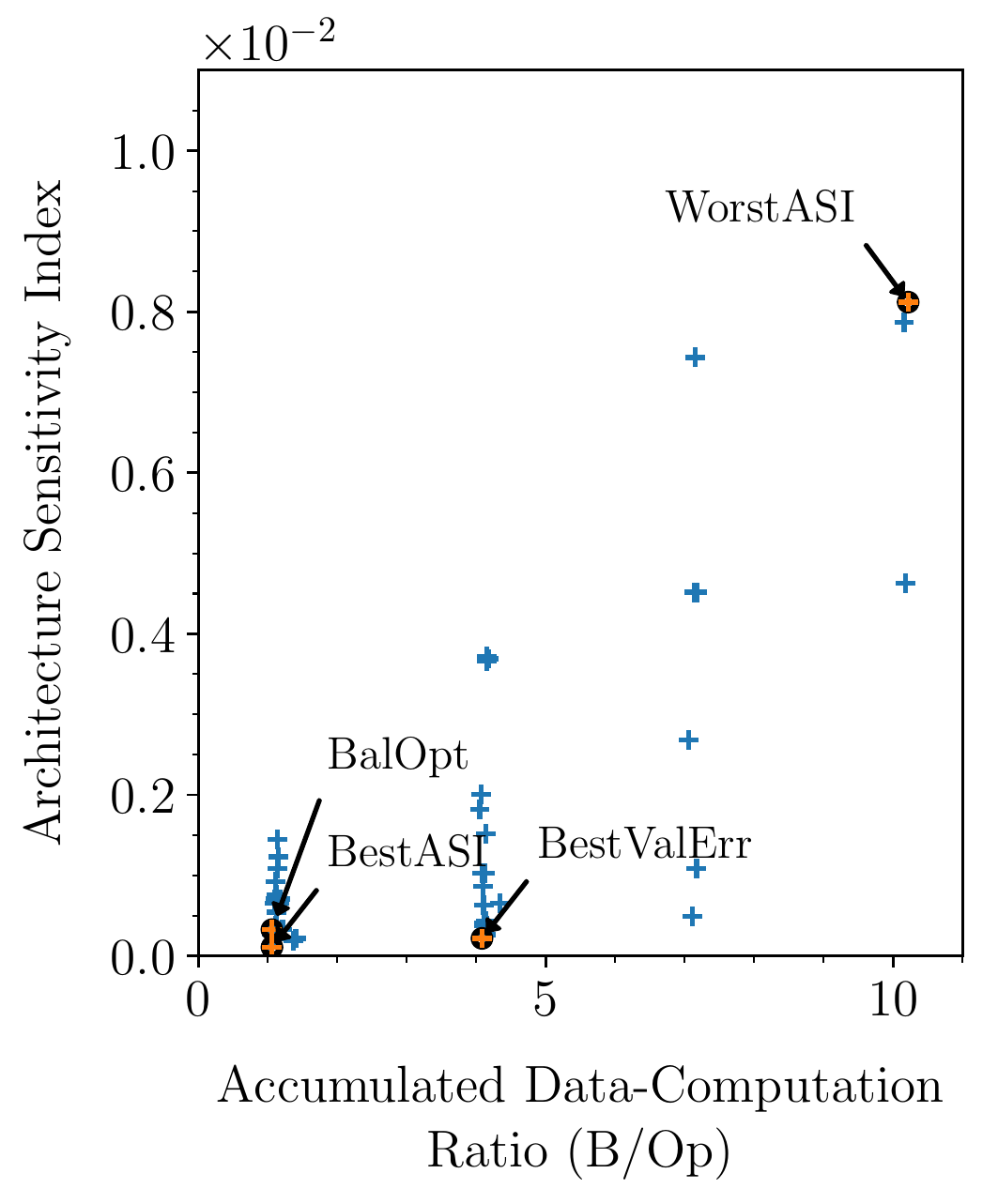}}
	\caption{Pairwise comparison of ASI with each of the other objective function outcomes for 50 Pareto-optimal architectures on GTSRB.}
	\label{fig:moo_gtsrb_pareto}
\end{figure*}

The resulting distributions of objective values for all $50$ models that were selected after the optimization with \pnash are shown in \figurename~\ref{fig:moo_cifar10_pareto} and \figurename~\ref{fig:moo_gtsrb_pareto} for CIFAR-10 and \ac{GTSRB}, respectively. The sub-figures (a)--(d) each depict the outcomes of $f_\mathrm{ASI}(N)$ versus one of the other objective functions. It can be seen that the WorstASI models have comparatively few operations and data transfers. However, the reverse is not always true, since there are models with few operations and data transfers as well as low \ac{ASI}. In other words, it is possible to have high efficiency and high error resilience at the same time.

\begin{figure}[!ht]
	\centering
	\subfloat[][CIFAR-10]{\includegraphics[width=0.23\textwidth]{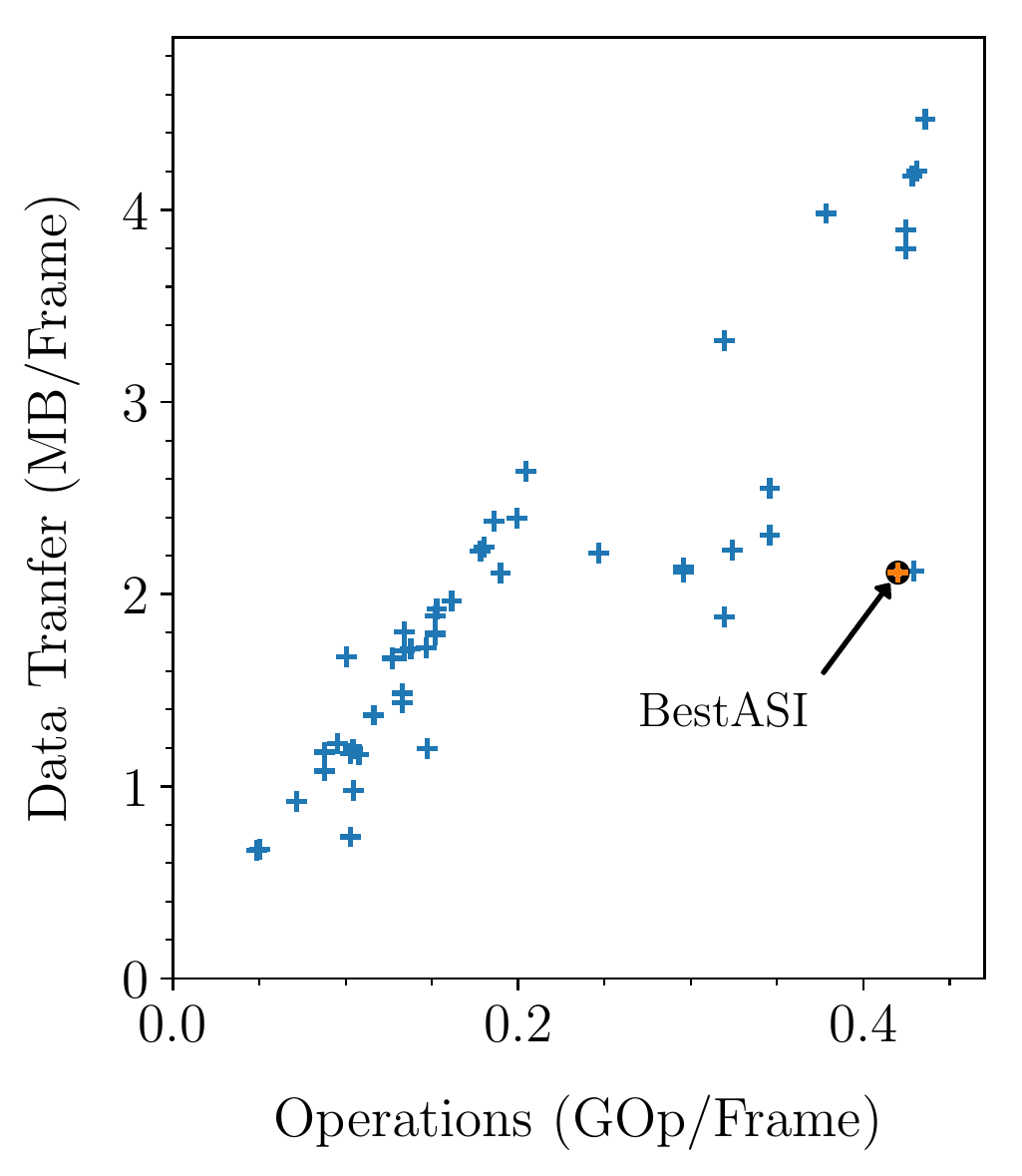}\label{fig:moo_data_vs_mac_cifar10}}
	\quad
	\subfloat[][GTSRB]{\includegraphics[width=0.23\textwidth]{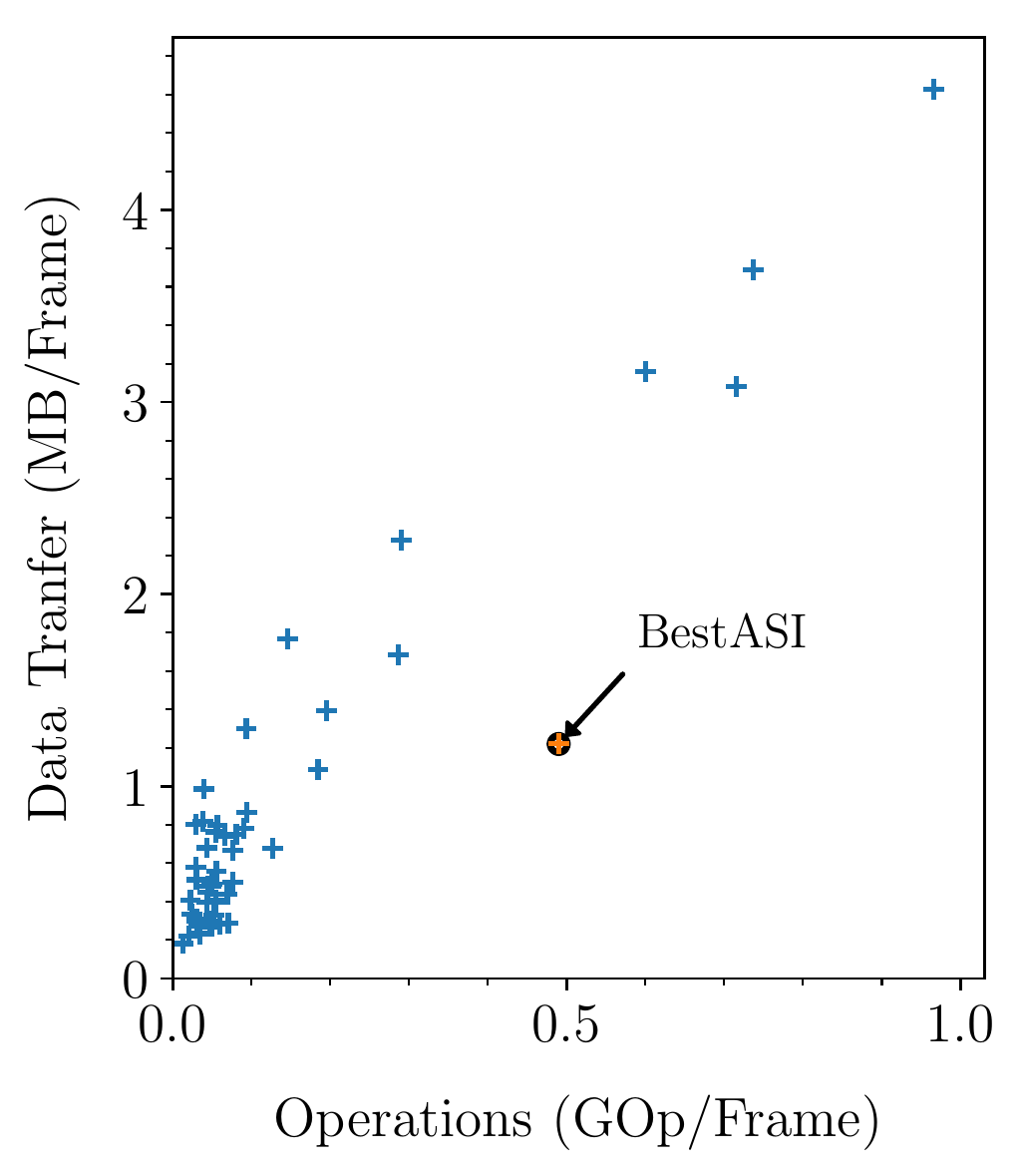}\label{fig:moo_data_vs_mac_gtsrb}}
	\caption{Data transfer vs. number of operations for Pareto-optimal architectures. BestASI models are located offside the main trend.}
	\label{fig:moo_data_vs_mac}
\end{figure}

Another interesting aspect visible in \figurename~\ref{fig:moo_cifar10_pareto}~(d) and \figurename~\ref{fig:moo_gtsrb_pareto}~(d) is a correlation between \ac{ADCR} and \ac{ASI}. Consequently, a low ratio of data transfers to operations is not only beneficial for limiting the required bandwidth of the \ac{DNN} accelerator, but also helps to reduce error sensitivity. This aspect becomes also apparent in \figurename~\ref{fig:moo_data_vs_mac}. It can be seen that models with more operations typically also require more data transfers. However, the BestASI models have a relatively high number of operations in comparison to their data transfers, as they are located offside the main trend in the scatter plot.

\subsubsection{Evaluation of resilience prediction}
We now evaluate the predictive performance of our \ac{ASI} metric by performing bit-flip fault injections using the framework described in Section~\ref{sec:fi_framework}. Bit-flips are randomly injected in all convolutional layer feature map outputs (after \ac{ReLU} activation and pooling, where applicable) that are written to memory. MinPQE quantization with 8 bits is used, except where otherwise specified. The value of each bit in the feature map outputs is toggled with a probability given by a defined \ac{BER}. To get statistically meaningful results \cite{Leve09}, random fault locations are sampled $n=200$ times and for each trial the effect on the classification output of the network is measured using the complete test set of the respective benchmark. For this purpose, the \ac{CCR}, \ie the fraction of images in the test set that are classified differently after the fault injection, is calculated. The sample mean of \ac{CCR} over all $n=200$ trials is reported. This can be interpreted as expected probability of \ac{SDC} at the given \ac{BER}.

\begin{figure}[!ht]
	\centering
	\subfloat[][CIFAR-10]{\includegraphics[width=0.23\textwidth]{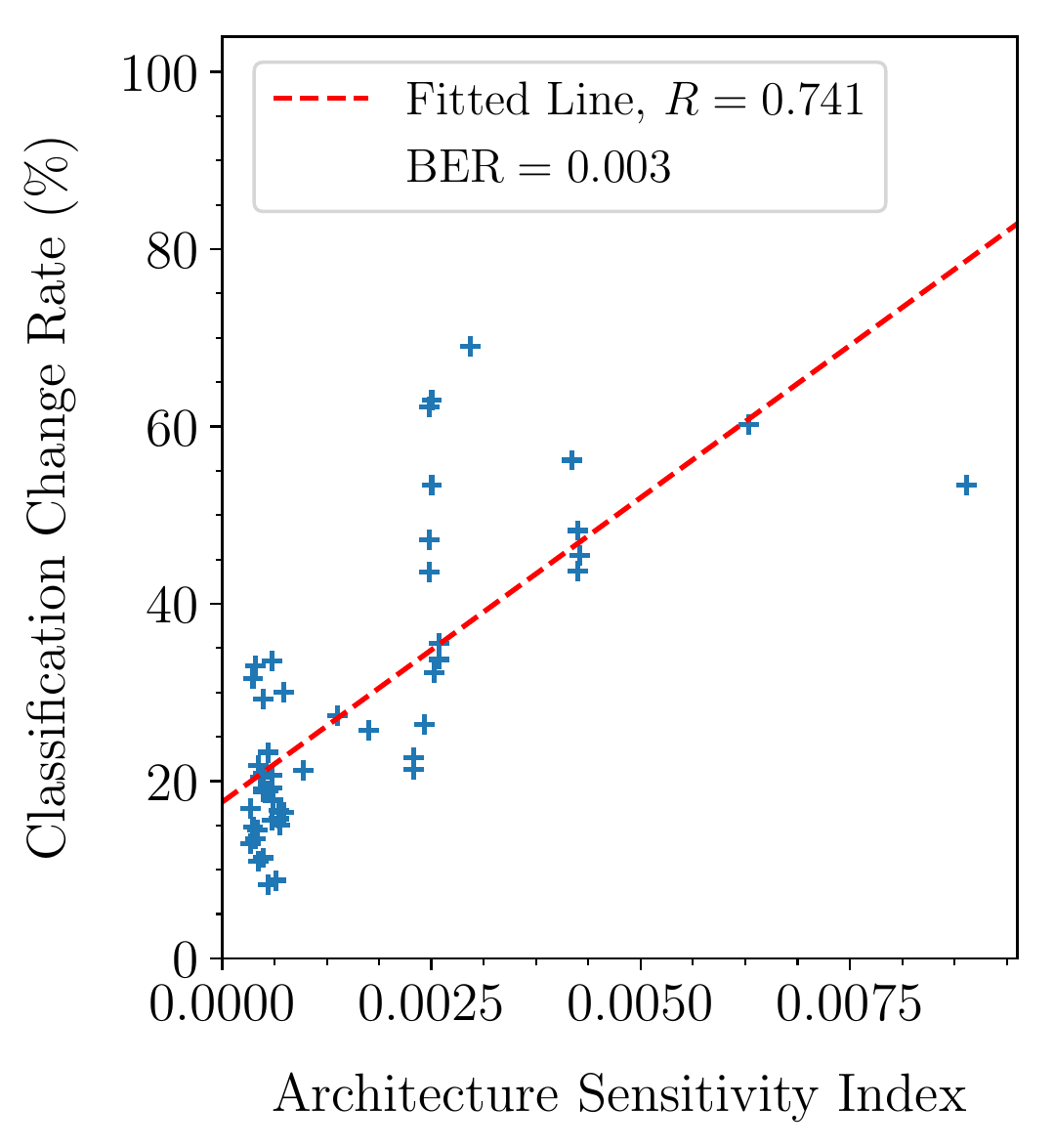}}
	\quad
	\subfloat[][GTSRB]{\includegraphics[width=0.23\textwidth]{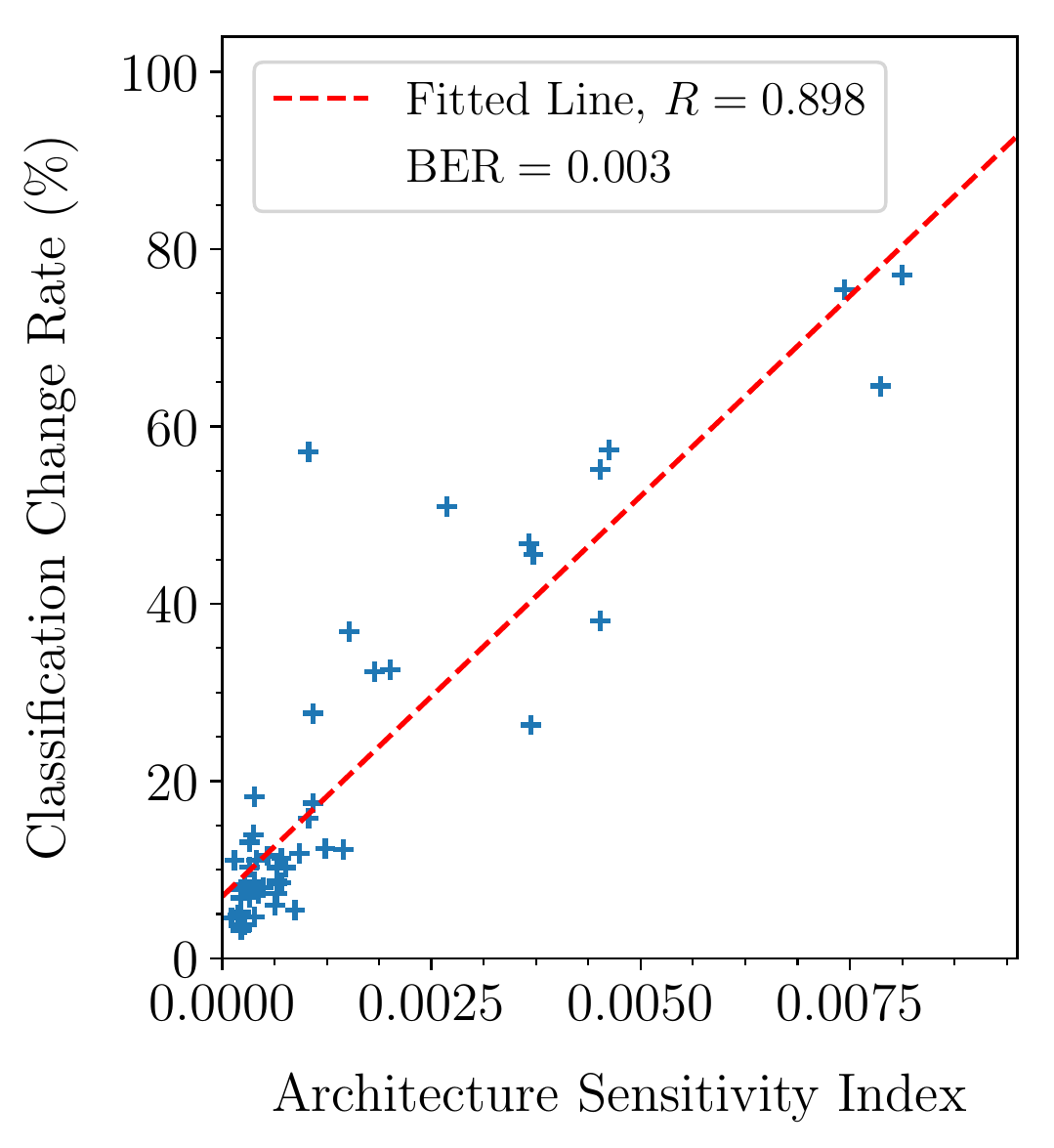}}
	\caption{Correlation of ASI and CCR. A correlation coefficient $R = 0.741$ is achieved for CIFAR-10 and $R = 0.898$ for \ac{GTSRB}.}
	\label{fig:moo_corr_asi}
\end{figure}

The results of a linear least-squares regression on the \ac{ASI} versus \ac{CCR} value pairs of the $50$ optimized models for each benchmark are shown in \figurename{}~\ref{fig:moo_corr_asi}. A \ac{BER} of 0.003 was used for bit-flip injections. A correlation coefficient $R = 0.741$ is achieved for CIFAR-10 and $R = 0.898$ for \ac{GTSRB}. While this indicates that the prediction is not 100\% accurate, the correlation is relatively strong. This is especially surprising, considering the fact that \ac{ASI} is completely determined by the architecture of the neural network and does not require any cumbersome measurements based on test data or weight parameters. Thus, we argue that \ac{ASI} is an efficient and useful metric to guide \ac{NAS} towards more resilient \ac{DNN} architectures.

\begin{figure}[!ht]
	\centerline{\includegraphics[width=0.48\textwidth]{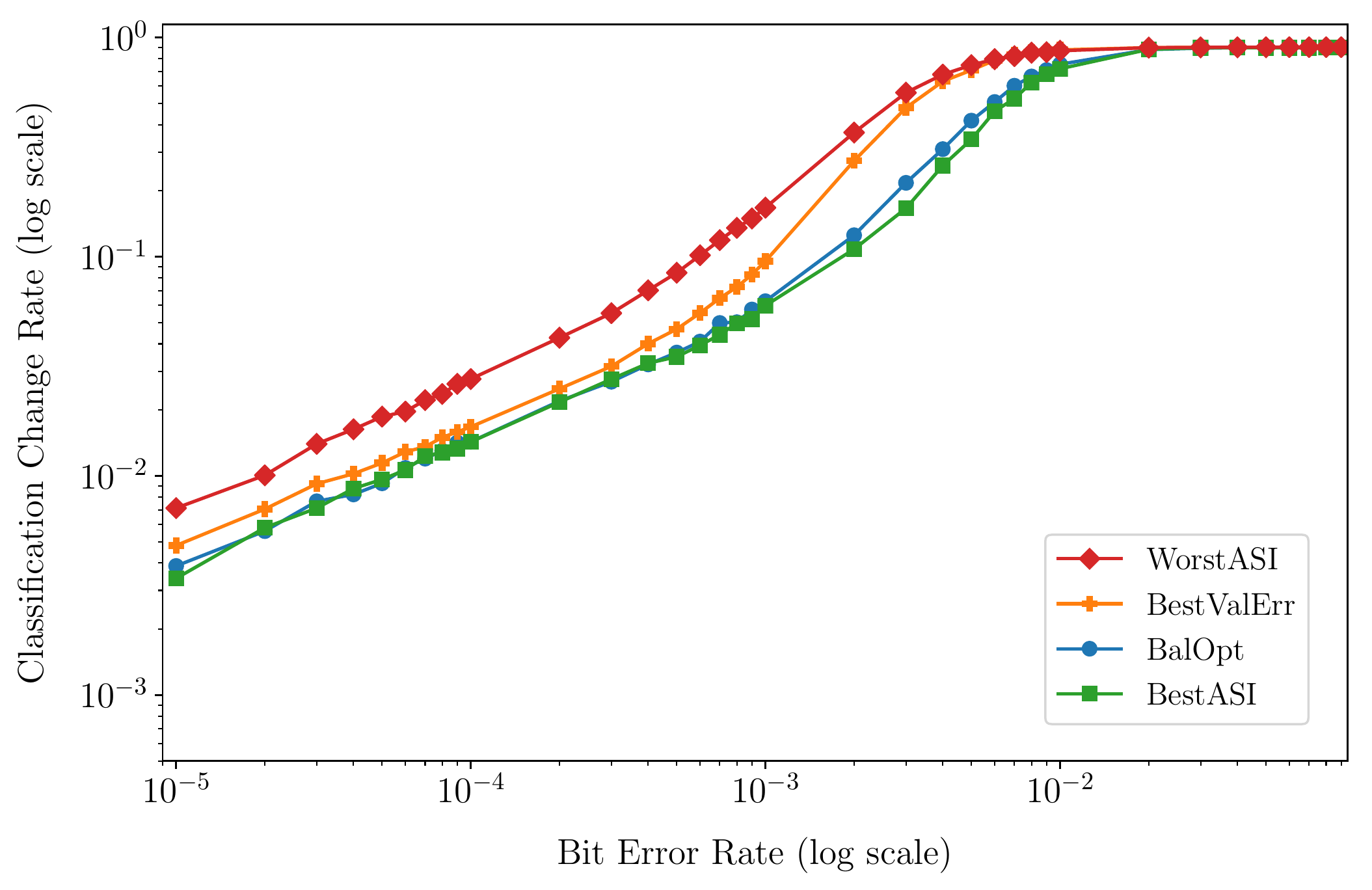}}
	\caption{Resulting CCR for different obtained optimizers on CIFAR-10 over a range of BERs.}
	\label{fig:moo_cifar10_ber_sweep}
\end{figure}

\begin{figure}[!ht]
	\centerline{\includegraphics[width=0.48\textwidth]{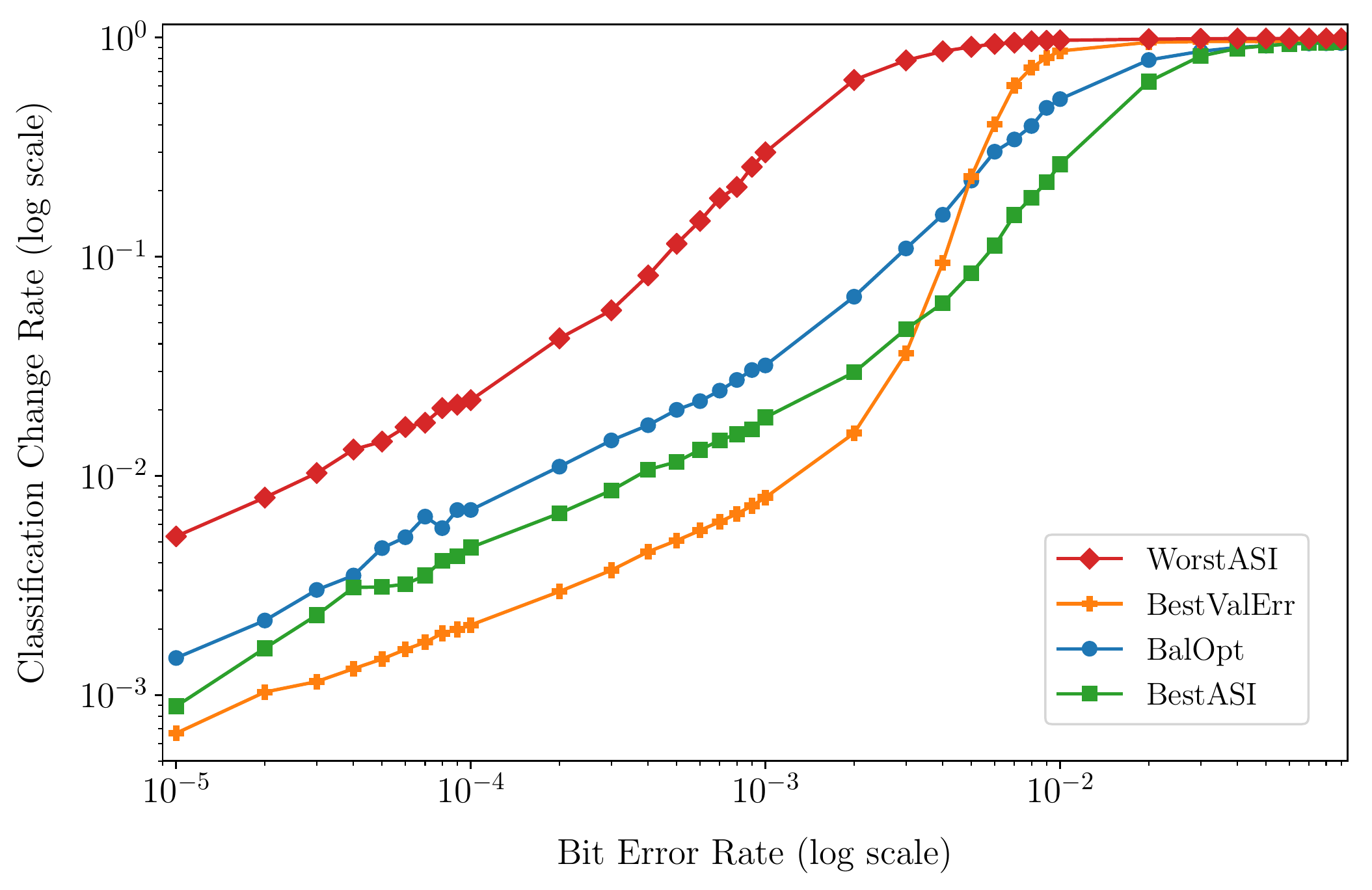}}
	\caption{Resulting CCR for different obtained optimizers on GTSRB over a range of BERs.}
	\label{fig:moo_gtsrb_ber_sweep}
\end{figure}

We also evaluate \acp{CCR} for varying \acp{BER} for a subset of models. The results for CIFAR-10 and \ac{GTSRB} are plotted in \figurename{}~\ref{fig:moo_cifar10_ber_sweep} and \figurename{}~\ref{fig:moo_gtsrb_ber_sweep}, respectively. An approximately linear dependency between \ac{BER} and \ac{CCR} can be observed at very low \aclp{BER}. At higher \acp{BER} a transition first to a rapid growth of \ac{CCR} (note the log scales) is visible and then the value saturates at a value corresponding to chance probability of choosing the same label after fault injection.

An interesting finding observable in \figurename{}~\ref{fig:moo_cifar10_ber_sweep} and \figurename{}~\ref{fig:moo_gtsrb_ber_sweep} is that the BestValErr models exhibit an unexpectedly low \ac{CCR} at low \acp{BER}, while they degrade less gracefully (much steeper increase \ac{CCR}) at high \acp{BER}. In the case of \ac{GTSRB} BestValErr is actually, despite its higher \ac{ASI}, much more resilient than BestASI at low \acp{BER}. An explanation might be that a good baseline classification performance adds an extra degree of error resilience, which is not captured by \ac{ASI}. The steeper increase, on the other hand, could be due to an overfitting to the task (\ie weaker ability for generalization).

\subsubsection{Comparison of quantization methods}

\begin{figure}[!ht]
	\centerline{\includegraphics[width=0.48\textwidth]{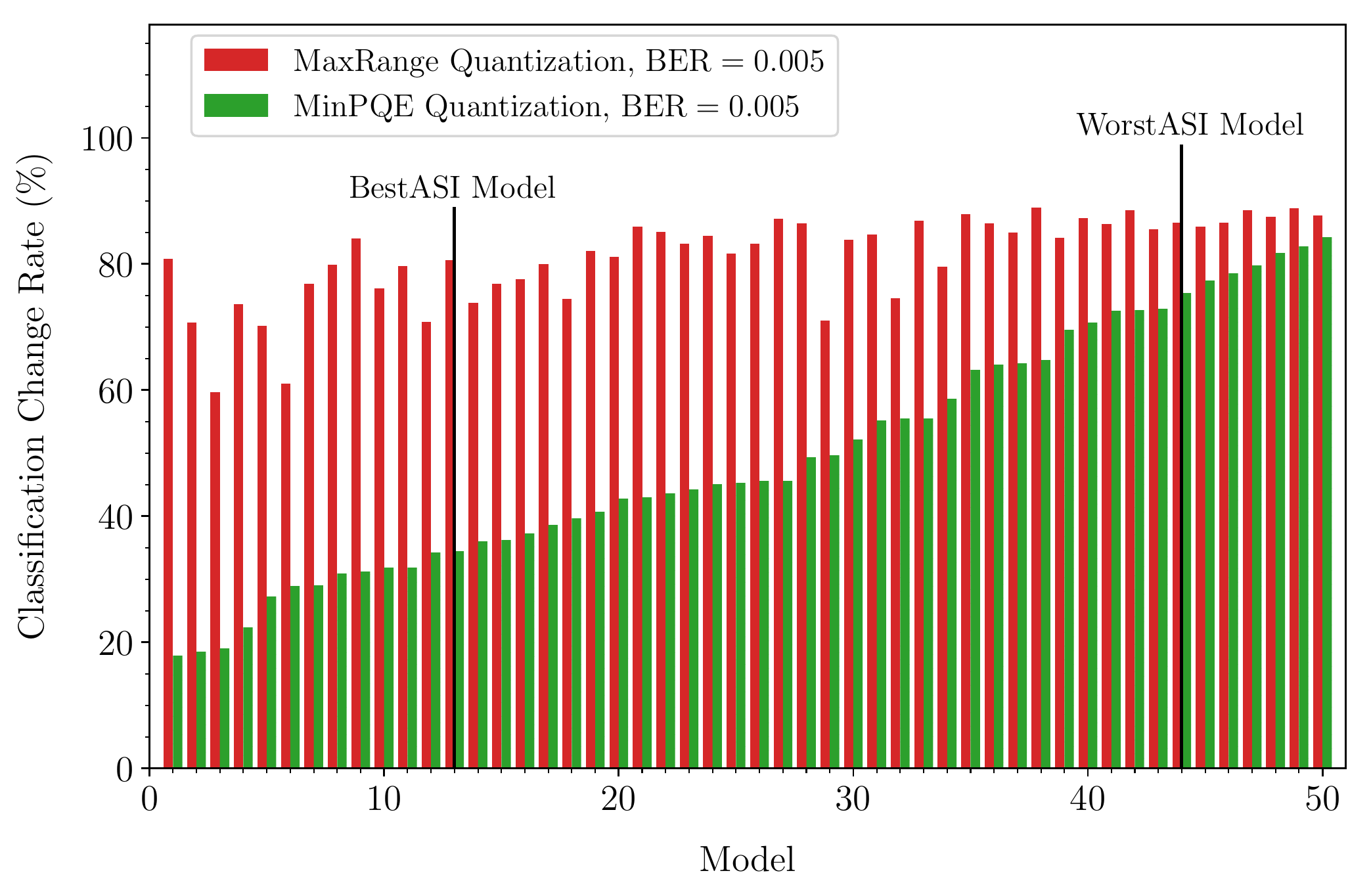}}
	\caption{Comparison of CCR at bit error rate 0.005 for CIFAR-10 models quantized with the MaxRange and MinPQE quantization methods. Models sorted after CCR observed with MinPQE quantization.}
	\label{fig:moo_cifar10_quant}
\end{figure}

\begin{figure}[!ht]
	\centerline{\includegraphics[width=0.48\textwidth]{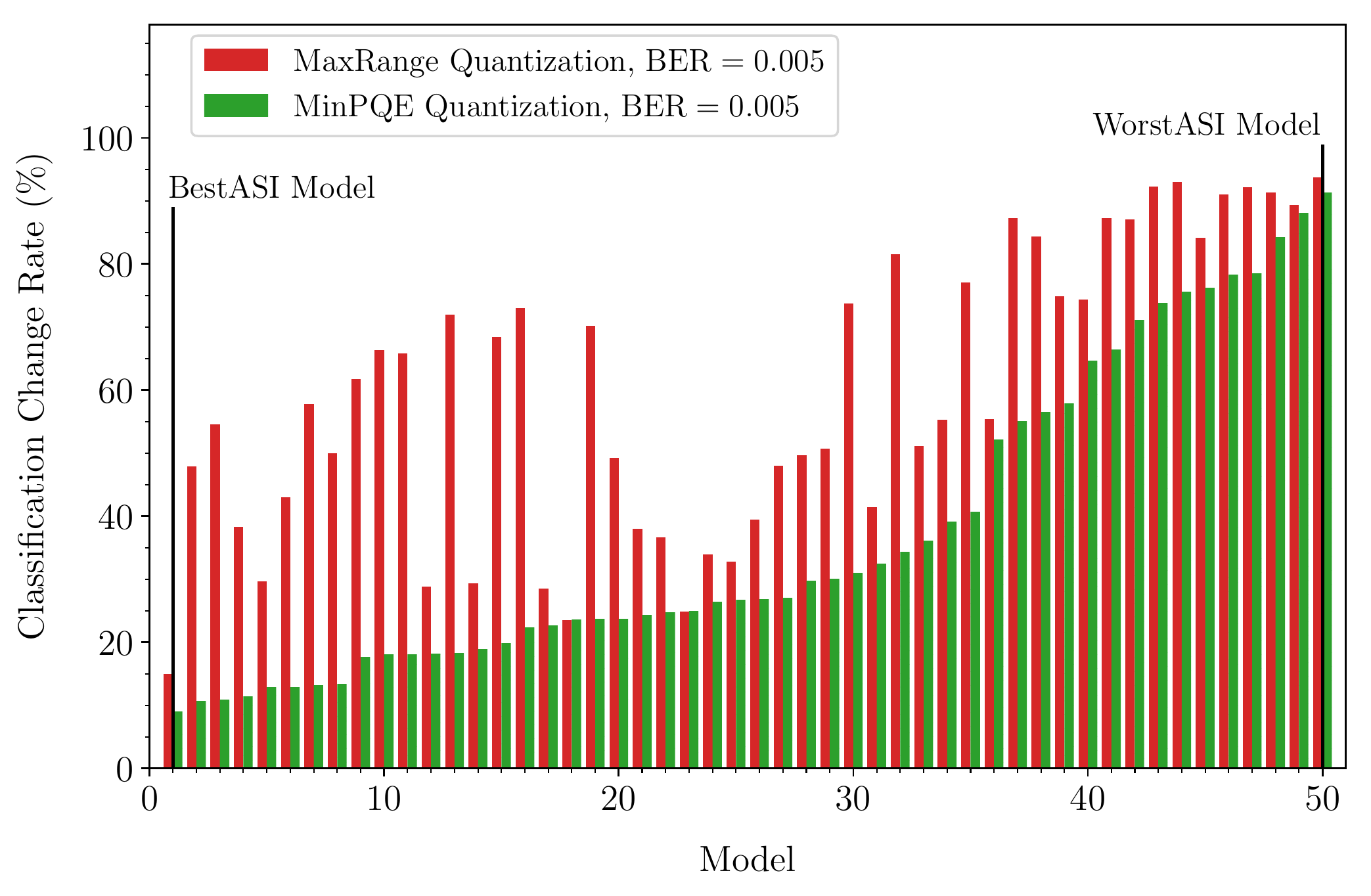}}
	\caption{Comparison of CCR at bit error rate 0.005 for GTSRB models quantized with the MaxRange and MinPQE quantization methods. Models sorted after CCR observed with MinPQE quantization.}
	\label{fig:moo_gtsrb_quant}
\end{figure}

Finally, we compare the MaxRange and MinPQE quantization methods (see Section~\ref{sec:quant_methods}), with respect to resulting \acp{CCR} after bit-flip fault injections with a \ac{BER} of 0.005. Results are shown in \figurename{}~\ref{fig:moo_cifar10_quant} and \figurename{}~\ref{fig:moo_gtsrb_quant}. The models are sorted in ascending order of \ac{CCR} after MinPQE quantization in these figures.

It can be seen that MaxRange results in a significantly worse \ac{CCR} in most of the cases. This can be explained by the fact that MaxRange tends to quantize values to a larger range, which is determined by far outliers, while these outliers are ignored (\ie clipped) by MinPQE. Consequently, MaxRange leads to a weaker \emph{signal-to-noise ratio} compared to MinPQE in the case of bit-flip errors. We thus argue that MinPQE is the preferable method, since it achieves both, low baseline classification error rates as well as high error resilience.

\section{Conclusions}
\label{sec:conclusions}

We have introduced a method for hardware-focused and automated neural architecture design. Our proposed hardware-specific objective functions, which only require network topology information for their evaluation, enable a fast design space exploration and finding of Pareto-optimal solutions of the \ac{NAS} algorithm. This makes our method efficient and applicable also for more complex classification benchmarks than the ones considered in this paper. We verified the accuracy of resilience prediction with memory bit-flip simulations and found it to be reasonably accurate to guide our \ac{NAS} algorithm towards architectural resilience optimization. Joint resilience, efficiency, and performance optimization has not been considered in the context of \ac{NAS} before. Finally, our findings about the influence of different quantization techniques on \ac{DNN} error resilience highlight the importance of choosing an optimization technique that fosters a high signal-to-noise ratio to limit the influence of bit-flip errors.


\bibliographystyle{spbasic}

\end{document}